%% file: main.tex
\documentclass[runningheads]{llncs}
\usepackage[T1]{fontenc}
\usepackage{graphicx}
\usepackage{booktabs}
\usepackage[misc]{ifsym}
\newcommand{\corr}{(\Letter)}
\usepackage{amsmath,amsfonts,amssymb}

\usepackage{xcolor}
\usepackage{hyperref}
\usepackage[capitalise]{cleveref}
\usepackage{subcaption}

\usepackage{tabularx}
\usepackage{multirow}
\usepackage{xspace}
\usepackage{verbatim}

\newcommand{\AUC}{$\mathrm{AULC}$\xspace}

\newcommand{\ie}{i. e.\xspace}
\newcommand{\eg}{e. g.\xspace}

\Crefname{equation}{Eq.}{Eqs.}
\Crefname{figure}{Fig.}{Figs.}
\Crefname{table}{Tab.}{Tabs.}
\Crefname{section}{Sec.}{Secs.}

\begin{document}

\title{Mind the Gap: A Framework for Assessing Pitfalls in Multimodal Active Learning}

\titlerunning{Mind the Gap: Pitfalls in Multimodal Active Learning}

\author{Dustin Eisenhardt\inst{1,2,3}\orcidID{0009-0007-6777-9038} \corr \and\\ Yunhee Jeong\inst{6}\orcidID{0000-0002-1150-7734} \and\\ Florian Buettner\inst{1,2,3,4,5}\orcidID{0000-0001-5587-6761}}



\institute{
    German Cancer Research Center (DKFZ), Heidelberg, Germany\and
    German Cancer Consortium (DKTK), partner site Frankfurt/Mainz, a partnership between DKFZ and UCT Frankfurt-Marburg, Frankfurt am Main, Germany\and
    Institute of Informatics, Goethe University Frankfurt, Frankfurt am Main, Germany\and
    Department of Medicine, Goethe University Frankfurt, Frankfurt am Main, Germany\and
    Frankfurt Cancer Institute  (FCI), Frankfurt am Main, Germany\and
    Crop Science Division, Bayer AG, Frankfurt am Main, Germany\\
    \email{dustin.eisenhardt@dkfz-heidelberg.de}
}

\maketitle              

\input{sections/00_abstract}

\input{sections/01_introduction}
\input{sections/05_sota}
\input{sections/20_pitfalls}
\input{sections/30_benchmark}
\input{sections/40_protocol}
\input{sections/50_evaluation}

\input{sections/80_discussion}
\input{sections/90_conclusion}
%
%
%
\bibliographystyle{splncs04}
\bibliography{main}
%

\clearpage
\appendix
\input{sections/93_app_params}
\input{sections/94_app_eval}
\end{document}

%% file: sections/00_abstract.tex
\begin{abstract}
Multimodal learning enables neural networks to integrate information from heterogeneous sources, but active learning in this setting faces distinct challenges.
These include missing modalities, differences in modality difficulty, and varying interaction structures.
These are issues absent in the unimodal case. 
While the behavior of active learning strategies in unimodal settings is well characterized, their behavior under such multimodal conditions remains poorly understood.

We introduce a new framework for benchmarking multimodal active learning that isolates these pitfalls using synthetic datasets, allowing systematic evaluation without confounding noise.
Using this framework, we compare unimodal and multimodal query strategies and validate our findings on two real-world datasets.

Our results show that models consistently develop imbalanced representations, relying primarily on one modality while neglecting others. 
Existing query methods do not mitigate this effect, and multimodal strategies do not consistently outperform unimodal ones. 
These findings highlight limitations of current active learning methods and underline the need for modality-aware query strategies that explicitly address these pitfalls.

Code and benchmark resources will be made publicly available.
\keywords{Multimodal Learning \and Active Learning.}
\end{abstract}

%% file: sections/01_introduction.tex
\section{Introduction}

Training modern neural networks requires large labeled datasets. 
When models must process inputs from multiple sources such as images, text, tabular features, or time-series signals, the annotation burden grows further, since each sample may require labeling across several modalities. 
This is a particular challenge for biomedical application, where labeling can only be performed by domain experts at high cost.
Active learning (AL) aims to reduce these annotation costs by selecting the most informative samples for labeling. 
Starting from a small labeled dataset, a model is trained and used together with a query strategy to iteratively select new samples for annotation until satisfactory performance is reached.

While AL has been studied extensively in unimodal settings, its behavior in multimodal learning remains not well understood. 
Multimodal models face several challenges that can hinder their ability to exploit information from all available modalities. 
In this work, we identify three such challenges, which we term pitfalls, and assess their implications for AL. 
First, modalities may be missing during training or inference.
Second, modality imbalance may arise when one modality contains signals that are easier to learn than those in another modality. 
Third, the usefulness of modalities depends on their interaction structure, \ie, whether information is shared across modalities, unique to one modality, or complementary such that both modalities are required for correct predictions.

These pitfalls can interact with and distort the behavior of AL methods. 
For example, if one modality is easier to learn, a model may rely on it primarily and develop impoverished representations of the remaining modalities.
As a result, the learned representations become imbalanced, with the model depending heavily on a single modality. 
This raises the question of whether existing query strategies can still identify informative samples when such imbalances are present.

To study this question, we introduce a framework for systematic benchmarking of multimodal AL that enables meaningful
statements about the performance of AL methods in multimodal settings. 
Our framework comprises a collection of synthetic datasets that isolate the proposed pitfalls and allow targeted analysis of their effects. 
By controlling modality missingness, difficulty and interaction structure, our framework enables systematic evaluation without confounding noise that is typically present in real-world datasets. 
In addition, we validate our findings on selected real-world datasets. 
Using this framework, we conduct an extensive empirical comparison of unimodal and multimodal AL query methods

Our experiments show that all pitfalls strongly affect learning behavior. 
In particular, we consistently observe that the network predominantly relies on one modality while largely ignoring the other.
Existing query methods are not able to mitigate this effect. 
Moreover, multimodal query strategies do not consistently outperform their unimodal counterparts in our experiments. 
These results reveal important limitations of current AL methods when applied in multimodal settings.

Our contributions are as follows:
\begin{itemize}
    \item We identify and formalize three key pitfalls in multimodal AL: missing modalities, modality imbalance, and varying modality interaction structures.
    \item We introduce a benchmarking framework with controlled synthetic datasets that systematically isolate each pitfall, enabling targeted analysis of its effect on multimodal AL performance.
    \item We provide the first comprehensive benchmark of multimodal AL, comparing unimodal and multimodal query strategies on both synthetic and real-world datasets and showing that existing methods fail to address the identified pitfalls.
\end{itemize}

%% file: sections/05_sota.tex
\section{Related Work}

\subsection{Active Learning}
\newcommand{\QRand}{\emph{Random}\xspace}
\newcommand{\QBADGE}{\emph{BADGE}\xspace}
\newcommand{\QBALD}{\emph{BALD}\xspace}
\newcommand{\QBMMAL}{\emph{BMMAL}\xspace}
\newcommand{\QEnt}{\emph{Entropy}\xspace}
\newcommand{\QGRACE}{\emph{GRACE}\xspace}
\newcommand{\QKCG}{\emph{KCG}\xspace}

AL aims to minimize the number of labeled samples required to train a well-performing model.
Starting from a small labeled budget and a large pool of unlabeled data, AL iteratively selects the most informative samples, has them labeled by an oracle, and adds them to the labeled set.

In AL, samples to be annotated are determined by a query method.
Querying strategies can be classified into three types: uncertainty-based, diversity-based and hybrid methods.
Uncertainty-based methods assume that samples with a high uncertainty add new knowledge to the training.
\QEnt acquires the samples with the highest entropy and \QBALD \cite{galDeepBayesianActive2017} chooses samples based on mutual information.
\QKCG \cite{senerActiveLearningConvolutional2018} is a diversity based method, which aims to cover the dataset as well as possible by solving the K-Center problem in the classifier's representation space.
Hybrid methods combine uncertainty and diversity sampling.
For example, \QBADGE\cite{ashDeepBatchActive2020} uses K-Means++ initialization to find cluster centroids in the gradient space of the classifier.

Several benchmarks exist for AL \cite{luthNavigatingPitfallsActive2023,wernerCrossDomainBenchmarkActive2024,jiRandomnessRootAll2023}, revealing important failure modes of unimodal query methods.
For example, many strategies fail to outperform random sampling under proper experimental controls \cite{luthNavigatingPitfallsActive2023}. 
However, none of these benchmarks account for challenges specific to multimodal data, such as modality imbalance or missing modalities. 
To the best of our knowledge, no benchmark for multimodal AL exists.

\subsection{Multimodal Active Learning}

In multimodal AL, samples contain inputs from multiple modalities.
\QBMMAL \cite{shenBalancedActiveLearning2023} and \QGRACE \cite{liGRACEGRadientbasedActive2024} are multimodal query methods that build on \QBADGE.
\QBMMAL introduces weights based on the contribution of each modality in order to achieve modality-balanced sample selection.
\QGRACE guides sample selection using measures of informativeness, representativeness and easiness scores that are adjusted over the course of AL cycles.

\cite{khanalMVAALMultimodalVariational2024} and \cite{rudovicMultimodalActiveLearning2019} modify the network architecture to learn to select informative samples, while \cite{shenEnhancingModalityRepresentation2024} align representations of different modalities and query samples with the smallest unimodal distances.
\cite{gordonFindRhinosFinding2023,zhangActiveLearningOpenSet2024,hassanCoherenceDrivenMultimodalSafety2025} propose query methods tailored to specific domains or tasks:
\cite{gordonFindRhinosFinding2023} use a domain-informed metric for wildlife monitoring, \cite{zhangActiveLearningOpenSet2024} design a query method specifically for object detection with vision-language models, and \cite{hassanCoherenceDrivenMultimodalSafety2025} leverage an external large language model for ranking samples.

\subsection{Multimodal Learning}

Multimodal learning aims to integrate information from heterogeneous sources to improve predictive performance. 

Missing modalities are a pervasive challenge in real-world multimodal systems, arising from sensor failures, privacy constraints, or data collection irregularities at any stage of a model's lifecycle \cite{wuDeepMultimodalLearning2025}. 
Existing approaches address this through composing or generating absent modalities from available ones, or through robust representation learning that maintains reliable predictions despite incomplete inputs \cite{wuDeepMultimodalLearning2025}.

A central challenge in multimodal learning is modality imbalance. Due to the simplicity bias of neural networks, multimodal models tend to rely on the modality that is easiest to learn, leaving other modalities under-trained even after convergence \cite{duModalityLazinessEverybodys2021,huangModalityCompetitionWhat2022}. 
Several remedies have been proposed, including gradient modulation strategies that dynamically balance modality contributions \cite{zhangMultimodalFusionLowquality2024}, and modality dropout \cite{neverovaModDropAdaptiveMultiModal2016}, which randomly drops modalities during training to preserve modality-specific representations while learning cross-modal correlations.

Beyond redundant information sharing, modalities can interact in richer ways: information may be unique to a single modality or synergistic, meaning it only becomes useful when both modalities are combined \cite{dufumierWhatAlignMultimodal2025}. 

%% file: sections/20_pitfalls.tex
\section{Multimodal Pitfalls in Active Learning}\label{sec:pitfalls}

In this section we present multimodal pitfalls and explain how they pose specific challenges in multimodal AL.
For clarity, we describe the pitfalls in a two-modality setting; they generalize straightforwardly to more modalities.

\subsubsection{P1: Missing Modalities}
Often it cannot be guaranteed that all samples contain every modality.
Instead, samples can be modality-incomplete at any stage of a model's life-cycle.
Training and inference have to work even if some samples have missing modalities.
This pitfall has different consequences depending on the stage of the neural network's life-cycle.

\noindent\underline{P1.1: Missing Modalities during Training} can contribute to insufficient learning of the modality that is missing more frequently.

\noindent\underline{P1.2: Missing Modalities during Inference} prevent neural networks from accessing the information from these modalities.

\noindent\emph{AL Relevance:}
The query method has to select samples so that the neural network is capable of learning all modalities adequately.
Modality missingness should be balanced and the creation of labeled sets that contain only a single modality should be avoided.

\subsubsection{P2: Modality Imbalance}
Neural networks tend to exploit the simplest learning pathways to solve a task, potentially causing the model to overlook the more challenging modality and fail to learn from it effectively.
Since different modalities come with varying degrees of complexity, \ie some being inherently easier to learn than others, it has been observed that multimodal neural networks overly rely on a single modality for prediction and neglect the rest. \cite{duModalityLazinessEverybodys2021}
We refer to the easier modality as the ‘strong’ modality. 
This imbalance causes insufficient representations to be learned for the weaker modality, degrading not only its individual performance but also the joint multimodal performance.

\noindent\emph{AL Relevance:}
Modality imbalance should not be reinforced by the query method. 
Sample acquisition should advance the joint-modality performance.
However, modality imbalance requires the query methods to choose samples that specifically contribute to the improvement of weak-modality as well.

\subsubsection{P3: Modality Interactions}
Information relevant for the task can be distributed over the modalities in different ways \cite{dufumierWhatAlignMultimodal2025}.

\noindent\underline{P3.1 Shared Information:} Information might be shared across modalities, so that each modality contains all information necessary to classify the target variable correctly.

\noindent\underline{P3.2 Unique Information:} In contrast, information might be unique to a single modality, \ie the task can be solved only when this modality is included.

\noindent\underline{P3.3 Complementary Information:} Information might be complementary in the modalities. In this case, the task is only solvable when the information from both modalities is aggregated.

\noindent\emph{AL Relevance:}
The informativeness of samples is certainly related to the way that information flows from the input modalities.
A query method should be able to reliably estimate sample informativeness no matter the type of modality interaction.

\subsubsection{Co-occurrence of Pitfalls}
These pitfalls are not mutually exclusive; in practice, they co-occur.
For example, in skin lesion classification, models may combine dermoscopic photographs with histopathology images obtained from biopsy slides. 
Histopathology images typically constitute the stronger modality due to their detailed cellular information (P2). 
However, their availability depends on the diagnostic stage: dermoscopic images are available during screening, whereas histopathology images are only obtained after a biopsy (P1). 
While some information is shared between modalities, \eg, that a lesion is abnormal, some information is only available via histopathology, \eg, the specific lesion subtype (P3). 
As a result, the classifier may struggle to make accurate predictions when only dermoscopic images are available, illustrating how modality imbalance and missing modalities can jointly degrade performance.

%% file: sections/30_benchmark.tex
\section{A Framework for Multimodal Active Learning}\label{sec:benchmark}

In this section, we explain the design of our benchmark.
Accordingly, we present a synthetic dataset that we use to create our experiments.
Furthermore, we present the experiments designed to evaluate the aforementioned pitfalls as well as the real-world datasets.

\subsection{QuintFeatures Dataset}\label{sec:quintfeatures}
\begin{figure}
    \centering
    \begin{subfigure}{0.3\textwidth}
        \includegraphics[width=\textwidth]{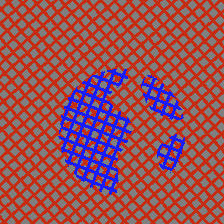}
        \caption{Blue Circle with Grid texture on Red Grid texture}
    \end{subfigure}
    \begin{subfigure}{0.3\textwidth}
        \includegraphics[width=\textwidth]{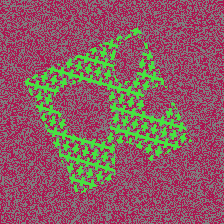}
        \caption{Green Square with Pulses texture on Purple Noise texture}
    \end{subfigure}
    \begin{subfigure}{0.3\textwidth}
        \includegraphics[width=\textwidth]{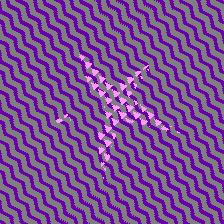}
        \caption{Pink Star with Triangles texture on Purple Zigzags texture}
    \end{subfigure}
    \caption{Examples of the QuintFeatures dataset.}
    \label{fig:quintfeatures}
\end{figure}

To rigorously evaluate our methods, we developed a novel synthetic dataset called QuintFeatures. 
It is inspired by Trifeatures\cite{hermannWhatShapesFeature2020}, but extends it through increasing feature dimensions and additional augmentation.
QuintFeatures generates realistic images by combining three distinct visual feature types in a controlled manner: shape, color, and texture.
Specifically, we randomly sample foreground shapes, colors, and textures alongside background colors and textures to create a diverse dataset suitable for testing performance under various conditions.
Shape, color and texture have ten different manifestations each.
In order to increase the difficulty of the dataset, we further add two types of augmentation.
First, we employ Perlin noise to erase parts of the foreground object.
Second, we randomize color while still keeping a clear distinction between different manifestations.
Examples are shown in \cref{fig:quintfeatures}.

\subsection{Pitfalls Coverage}\label{sec:benchmark-pitfalls}
In this section, we explain how the pitfalls P1 - P3 are covered by our framework and introduce four experiments that allow for a systematic evaluation.

\subsubsection{P1.1 (Missing):}\label{sec:benchmark-pitfalls-missing}
For evaluating missingness during training we construct a dataset, called \emph{Missing}, that contains two input modalities that are missing with different probabilities.
The goal of this experiment is to predict the target variable without neglecting the first modality, which is missing more often.
Each modality is sampled from the QuintFeatures dataset (\cref{sec:quintfeatures}).
The target task is to predict the foreground shape of the images. 
The images are sampled such that the shape is identical between both modalities, representing completely shared information.
Other features are randomized.
In order to model the pitfall, we set a missing modality probability of 90\% for the first and 10\% for the second modality. 
If at any step two modalities would be missing, we randomly choose one to keep.

\subsubsection{P2 \& P3.1 (Share):}\label{sec:benchmark-pitfalls-share}
We create a setting, called \emph{Share}, where both modalities contain all information needed to solve the target task, yet differ in complexity. 
The goal is to assess whether the model can learn to exploit the weaker modality despite the availability of an easier one. 
Because the features of the QuintFeatures dataset are of similar complexity, we instead construct a synthetic bimodal dataset by treating CIFAR-10 \cite{krizhevskyLearningMultipleLayers} and MNIST \cite{MNISTHandwrittenDigit2020} as separate modalities. 
To ensure shared information, each class from CIFAR-10 is mapped to exactly one class from MNIST.

\subsubsection{P3.2 (Unique):}\label{sec:benchmark-pitfalls-unique}
In this experiment, called \emph{Unique}, we evaluate the case where some information is only available to one modality.
The goal of this experiment is to predict the power-set of the shared and unique features, while maximizing the performance of each modality.
Therefore, we construct a dataset containing two input modalities.
Again, each modality is sampled from the QuintFeatures dataset (\cref{sec:quintfeatures}).
The target task is to predict the power-set of both features shapes and the texture of the first modality.
The images are sampled such that the shape is identical between both modalities.
Since the second modality does not have access to the texture information portrayed in the first modality, a maximum accuracy of 10\% is achievable. 

\subsubsection{P3.3 (Synergy):}\label{sec:benchmark-pitfalls-synergy}
In this experiment, called \emph{Synergy}, we assess the case where information between modalities is complementary, \ie, useful for prediction only when both modalities are present. 
The goal is to evaluate whether models and query strategies can leverage synergies between the modalities, maximizing joint prediction performance. 
We construct a bimodal dataset using QuintFeatures (\cref{sec:quintfeatures}), where the target task is to predict the class in the power-set of the first modality’s texture and the second modality’s shape.

\subsubsection{P1.2:}\label{sec:benchmark-pitfalls-evaluation}

We conduct an exhaustive evaluation of all combinations of missing modalities in all experiments (\cref{sec:systematic-evaluation}).
For example, given two modalities A and B, we evaluate the model on the test set three times: once keeping only modality A, once keeping only modality B and once with both modalities.
Thereby, we test the model's capabilities in any possible setting of missing modalities during inference time.

\subsection{Real-World Datasets}

To assess the generalizability of our findings, we utilize two real-world datasets that mirror the share setting but represent diverse modalities. 
These datasets serve as a validation of our framework, demonstrating whether the identified pitfalls and query method performance observed with synthetic data also hold true in practical scenarios.

\subsubsection{Food101}
UPMC Food101\cite{wangRecipeRecognitionLarge2015} is a large multimodal classification dataset that contains about 100,000 samples for 101 food categories.
Each sample consists of the text of one recipe and an image from the recipe’s webpage. 
Because the dataset was collected automatically, samples may contain misleading information
For example, samples labeled `hamburger' might contain images of hamburgers (food), burger patties (ingredients), or unrelated content such as a stew (noise).

\subsubsection{MIMIC-IV}
MIMIC-IV \cite{goldbergerPhysioBankPhysioToolkitPhysioNet2000,johnsonMIMICIVFreelyAccessible2023,PhysioNet-mimiciv-2.2,PhysioNet-mimic-iv-ecg-1.0} is a large database of electronic health records. 
We use the bloodlabs and electrocardiograms (ECGs) from MIMIC-IV in order to classify 1-year patient mortality, \ie whether the patient dies within one year past admission or not.
Further information can be found in \cref{sec:settings-mimic}.

%% file: sections/40_protocol.tex
\section{Experimental Setup}

We now describe the experimental setup used to evaluate multimodal AL under the pitfalls from \cref{sec:pitfalls}.

\subsubsection{Model Architecture}
For all experiments, we use a simple architecture that uses uni-modal encoders and a fusion layer.
The fusion layer combines the uni-modal features and provides a multimodal latent space.
A linear layer is applied on top of the multimodal features to classify the inputs.
More details on the model architecture can be found in \cref{sec:settings-arch}.

\subsubsection{Hyperparameter Selection}

For hyperparameter selection we follow the approach of \cite{luthNavigatingPitfallsActive2023}. 
We use a fixed training recipe specifying training epochs, batch size, optimizer, and learning rate scheduler (\cref{sec:settings-training}). 
The learning rate, weight decay, and augmentations are optimized using the starting budget and the validation set at the beginning of the AL cycle. 
Final hyperparameters are reported in  \cref{sec:settings-training}.

\subsubsection{Active Learning Settings}

In order to simulate a variety of conditions, we evaluate different label regimes on each dataset.
Specifically, a `low-label', `mid-label' and a `high-label' regime.
These affect the size of the starting budget, validation set and acquisition.
More details can be found in \cref{sec:settings-al}.

\subsubsection{Query Methods}

We include five unimodal query methods (\QBADGE, \QBALD, \QEnt, \QKCG, \QRand) following \cite{luthNavigatingPitfallsActive2023}, along with two multimodal query methods (\QBMMAL, \QGRACE).
We exclude other multimodal methods for the following reasons: 
\cite{khanalMVAALMultimodalVariational2024,rudovicMultimodalActiveLearning2019,shenEnhancingModalityRepresentation2024} require modifications to the training pipeline, either through architectural changes \cite{khanalMVAALMultimodalVariational2024,rudovicMultimodalActiveLearning2019} or additional loss terms \cite{shenEnhancingModalityRepresentation2024}, whereas our focus is on self-contained query methods that do not alter the model or training procedure. 
\cite{gordonFindRhinosFinding2023,zhangActiveLearningOpenSet2024,hassanCoherenceDrivenMultimodalSafety2025} are designed for specific application domains or tasks and are thus incompatible with our domain-agnostic classification setting.

\subsubsection{Modality Dropout}\label{sec:setup-moddrop}
To simulate a practical application of AL on multimodal data, where strategies to mitigate common multimodal pitfalls are often employed, we incorporate Modality Dropout (ModDrop) \cite{neverovaModDropAdaptiveMultiModal2016}. 
ModDrop randomly drops out single modalities during training.
This technique promises to preserve uniqueness of each modality-specific representation while learning cross-modal correlations.
All synthetic experiments are conducted with and without ModDrop.
For the real-world experiments, we ran preliminary experiments using the hyperparameters from \cref{sec:settings-training} and found that ModDrop is beneficial for Food101 and MIMIC.
Therefore, we use ModDrop for these datasets.

\subsubsection{Evaluation}
For each evaluated dataset, we report performance based on the Area Under the Learning Curve (AULC). 
This metric aggregates the balanced test accuracy of the best-performing models obtained after every AL iteration:

$$\mathrm{AULC} = \sum_{i=0}^{N-1}\frac{1}{2}\frac{\mathrm{ACC}(i+1)-\mathrm{ACC}(i)}{i+1-i}, \mathrm{ACC}(0)=0$$

\noindent where $\mathrm{ACC}(i)$ denotes the balanced test accuracy at AL iteration $i$ out of a total of $N$ iterations. 
The \AUC is bounded within the range  $[0, N]$.

We report this metric for all query methods across all three regimes.
Furthermore, in order to evaluate the pitfalls, we exhaustively assess model performance on every possible combination of missing modalities (\cref{sec:benchmark-pitfalls-evaluation}).
All results are derived from three experiments per setting and query method, varying seeds and training/validation splits while keeping the test set constant across runs.

%% file: sections/50_evaluation.tex
\begin{table}[t]
\caption{Random Query Method on Synthetic Data. \AUC for the `high-label' regime using the \emph{Random} query method, evaluated on each modality combination separately. $^+$: Scores of partially predictive modalities are scaled by a factor of ten for comparability.}\label{tab:results-pitfalls}

\centering
\begin{tabularx}{\textwidth}{XXXXX}
\toprule
\multicolumn{1}{l}{\textbf{Dataset}} & \multicolumn{1}{l}{\textbf{ModDrop}} & \multicolumn{3}{c}{\textbf{AULC}}\\
&&\multicolumn{1}{c}{Modality A} & \multicolumn{1}{c}{Both} & \multicolumn{1}{c}{Modality B} \\
\midrule
Missing && Rare & Both & Frequent \\\cmidrule{3-5}
                                    & Yes                                     & 1.01 ± 0.08                    & 7.94 ± 0.10                         & 7.95 ± 0.11                    \\
                                    & No                                      & 0.97 ± 0.05                    & 7.94 ± 0.13                         & 7.96 ± 0.13                    \\

Share && CIFAR-10 & Both & MNIST \\\cmidrule{3-5}
                                    & Yes                                     & 5.10 ± 0.11                    & 8.91 ± 0.01                         & 8.81 ± 0.02                    \\
                                    & No                                      & 4.23 ± 0.24                    & 8.91 ± 0.01                         & 8.82 ± 0.02                    \\

Unique && Full & Both & Partial$^+$ \\\cmidrule{3-5}
                                    & Yes                                     & 8.45 ± 0.07                    & 8.62 ± 0.09                         & 6.40 ± 0.96                    \\
                                    & No                                      & 8.45 ± 0.10                    & 8.51 ± 0.06                         & 1.76 ± 1.43                    \\

Synergy && Partial A$^+$ & Both & Partial B$^+$ \\\cmidrule{3-5}
                                    & Yes                                     & 8.86 ± 0.17                    & 2.49 ± 1.54                         & 2.54 ± 1.58                    \\
                                    & No                                      & 8.88 ± 0.09                    & 6.55 ± 1.19                         & 6.60 ± 1.22                    \\\bottomrule \\
\end{tabularx}

\end{table}

\section{Systematic Evaluation}\label{sec:systematic-evaluation}

We now present a systematic evaluation of multimodal AL under the pitfalls introduced in \cref{sec:pitfalls}. 
We first assess the effect of each pitfall on model performance in isolation (\cref{sec:pitfalls-effect}), then compare unimodal and multimodal query methods on synthetic (\cref{sec:results-synthetic}) data and verify our findings on real-world datasets (\cref{sec:results-rw}). 
Detailed plots of all results can be found in \Cref{sec:detailed-results}

\subsection{Effect of the Pitfalls}\label{sec:pitfalls-effect}

In this section we investigate the impact of the experiments proposed in \cref{sec:benchmark-pitfalls} on the learning performance of a multimodal neural network. 
\Cref{tab:results-pitfalls} shows the \AUC of the \QRand query method in the high-label regime of the synthetic datasets.
We further compare the effect of applying ModDrop (\cref{sec:setup-moddrop}).
The following paragraphs discuss the results one by one, based on the pitfalls.

\subsubsection{P1.1 (Missing)}
The \AUC achieved when evaluated only on the rare modality is significantly smaller than that obtained on the frequent one.
The joint \AUC is close to that achieved on the frequent modality only.
Employing ModDrop does not yield substantial differences.

\subsubsection{P2 \& P3.1 (Share)}
The \AUC when evaluated only on CIFAR-10 is considerably smaller than when only evaluated on MNIST.
The joint \AUC is close to the \AUC achieved on MNIST only.
Here, applying ModDrop increases the performance on the weak modality while leaving the strong modality unaffected.
Though, a significant gap between the weak and the strong modalities' performances remains.

\subsubsection{P3.2 (Unique)}
The modality with all task-relevant information performs substantially better than the one with only partial information.
The joint \AUC is similar to that of the modality with full information.
In this case, ModDrop significantly improves the \AUC gained in the case of partial task-information.
The other modality remains unaffected and a performance gap between both modalities remains.

\subsubsection{P3.3 (Synergy)}
The model performs best on modality A, achieving considerably higher scores than using modality B.
In contrast to previous situations, the joint \AUC is close to the weak modalities performance.
Here, ModDrop harms the overall performance, resulting in significantly lower scores.

\subsubsection{Summary}
Across all four experimental settings, modality imbalance consistently emerges: models disproportionately rely on one modality regardless of the information structure. 
ModDrop mitigates this imbalance in all but the synergy setting, where it is actively harmful, reducing joint performance substantially. 
Based on these results, further evaluation of the synthetic datasets will use ModDrop for all but the synergy dataset.

\begin{table}[p]
\caption{Query Method Comparison on Synthetic Data. \AUC per query method on synthetic datasets across all label regimes. Results are reported separately for each modality combination. Best entries are highlighted in \textbf{bold}, the weak modality in \textit{italics}.}\label{tab:results-synthetic}

\centering
\begin{tabularx}{\textwidth}{llXccccccc}
\toprule
\multicolumn{2}{c}{\multirow{2}{1.5cm}{\textbf{Dataset Regime}}} & \textbf{Metric} & \multicolumn{7}{c}{\textbf{AULC per Query Method}}\\\cmidrule{4-10}
                         & & & BADGE     & BALD      & BMMAL     & Entropy   & GRACE     & KCG       & Random    \\
\midrule

\multirow{9}{*}{\rotatebox{90}{Missing}} & \multirow{3}{*}{\rotatebox{90}{Low}}
 & \textit{Rare}    & 1.05  & 1.07  & 1.13  & \textbf{1.19}  & 1.10  & \textbf{1.19}    & 0.97 \\
&& Both             & \textbf{1.13}  & 1.02  & 1.06  & 1.12  & 1.02  & 1.22    & 0.96  \\
&& Frequent         & 1.08  & 0.91  & 0.96  & 0.94  & 0.91  & \textbf{1.10}    & 0.92  \\\cmidrule{3-10}

& \multirow{3}{*}{\rotatebox{90}{Mid}}
 & \textit{Rare}    & 1.50  & 2.33  & 2.46  & 3.49  & 2.91  & \textbf{4.99}    & 1.01 \\
&& Both             & 5.77  & 5.25  & 5.63  & 5.01  & 5.22  & \textbf{6.09}    & 5.51  \\
&& Frequent         & \textbf{5.71}  & 4.98  & 5.43  & 4.26  & 4.85  & 5.00    & 5.52  \\\cmidrule{3-10}

& \multirow{3}{*}{\rotatebox{90}{High}}
 & \textit{Rare}    & 1.13  & 3.79  & 2.19  & \textbf{5.45}  & 2.20  & 1.03    & 1.01 \\
&& Both             & 8.02  & 7.90  & 8.02  & \textbf{8.06}  & 7.96  & 7.96    & 7.94 \\
&& Frequent         & \textbf{8.03}  & 7.69  & 7.95  & 7.49  & 7.86  & 7.97    & 7.95 \\\midrule


\multirow{9}{*}{\rotatebox{90}{Share}} & \multirow{3}{*}{\rotatebox{90}{Low}}
 & \textit{CIFAR-10}    & \textbf{3.66}  & 3.52  & \textbf{3.66}  & 3.24  & 3.45  & 3.53    & \textbf{3.66}  \\
&& Both                 & \textbf{8.42}  & 8.32  & 8.41  & 7.99  & 8.36  & 8.30    & \textbf{8.42}  \\
&& MNIST                & \textbf{8.51}  & 8.46  & 8.49  & 8.19  & 8.48  & 8.45    & 8.49  \\\cmidrule{3-10}

& \multirow{3}{*}{\rotatebox{90}{Mid}}
 & \textit{CIFAR-10}    & 4.62  & 4.59  & 4.72  & 4.41  & 4.47  & 4.65    & \textbf{4.76}  \\
&& Both                 & 8.85  & 8.86  & \textbf{8.87}  & 8.64  & 8.81  & 8.80    & 8.86  \\
&& MNIST                & 8.78  & 8.78  & \textbf{8.80}  & 8.54  & 8.74  & 8.73    & 8.77  \\\cmidrule{3-10}

& \multirow{3}{*}{\rotatebox{90}{High}}
 & \textit{CIFAR-10}    & 4.91  & 4.94  & 5.08  & 4.78  & 4.80  & 5.00    & \textbf{5.10} \\
&& Both                 & 8.91  & 8.92  & \textbf{8.93}  & 8.85  & 8.90  & 8.91    & 8.91 \\
&& MNIST                & 8.83  & \textbf{8.84}  & \textbf{8.84}  & 8.75  & 8.81  & 8.83    & 8.81 \\\midrule


\multirow{9}{*}{\rotatebox{90}{Unique}} & \multirow{3}{*}{\rotatebox{90}{Low}}
 & Full                 & \textbf{4.75}  & 4.58  & 4.63  & 3.95  & 4.41  & 4.21    & 4.74  \\
&& Both                 & \textbf{4.80}  & 4.61  & 4.76  & 4.14  & 4.66  & 4.29    & 4.76  \\
&& \textit{Partial}     & 2.33  & 2.20  & 2.96  & 3.66  & \textbf{3.75}  & 2.74    & 1.94  \\\cmidrule{3-10}

& \multirow{3}{*}{\rotatebox{90}{Mid}}
 & Full                 & \textbf{6.48}  & 6.38  & 6.41  & 6.29  & 6.23  & 6.17    & 6.43  \\
&& Both                 & \textbf{6.56}  & 6.49  & 6.46  & 6.38  & 6.38  & 6.39    & 6.52  \\
&& \textit{Partial}     & 2.53  & 2.86  & 2.55  & 2.73  & 3.61  & \textbf{3.99}    & 2.87  \\\cmidrule{3-10}

& \multirow{3}{*}{\rotatebox{90}{High}}
 & Full                 & \textbf{8.47}  & 8.44  & 8.43  & 8.40  & 8.45  & 8.44    & 8.45 \\
&& Both                 & \textbf{8.64}  & 8.62  & 8.60  & 8.56  & 8.62  & 8.59    & 8.62 \\
&& \textit{Partial}     & 6.26  & 6.26  & 5.98  & 5.85  & 6.22  & 5.86    & \textbf{6.40} \\\midrule


\multirow{9}{*}{\rotatebox{90}{Synergy}} & \multirow{3}{*}{\rotatebox{90}{Low}}
 & Partial A            & 6.11  & 6.21  & 6.13  & 5.90  & 6.08  & 6.00    & \textbf{6.22}  \\
&& Both                 & 3.69  & 3.30  & 3.70  & 2.99  & 3.85  & \textbf{4.06}    & 3.72  \\
&& \textit{Partial B}   & 4.59  & 4.01  & 4.34  & 3.64  & 4.47  & 4.95    & \textbf{4.61}  \\\cmidrule{3-10}

& \multirow{3}{*}{\rotatebox{90}{Mid}}
 & Partial A            & \textbf{7.89}  & 7.84  & 7.84  & 7.76  & 7.74  & 7.87    & 7.81  \\
&& Both                 & 5.42  & 4.85  & 5.68  & 5.54  & 5.40  & \textbf{5.91}    & 5.53  \\
&& \textit{Partial B}   & 5.42  & 5.30  & 6.29  & 6.06  & 5.84  & \textbf{6.42}    & 5.90  \\\cmidrule{3-10}

& \multirow{3}{*}{\rotatebox{90}{High}}
 & Partial A            & 8.91  & 8.89  & 8.89  & 8.86  & \textbf{8.92}  & 8.91    & 8.88 \\
&& Both                 & \textbf{7.89}  & 7.76  & 7.50  & 7.04  & 7.85  & 7.35    & 6.55 \\
&& \textit{Partial B}   & \textbf{7.90}  & 7.79  & 7.55  & 7.09  & 7.89  & 7.40    & 6.60 \\\midrule

\bottomrule
\end{tabularx}
\end{table}

\subsection{Query Method Comparison on Synthetic Data}\label{sec:results-synthetic}

Having investigated the effects of the multimodal pitfalls, we now examine unimodal and multimodal query methods on the synthetic settings (\cref{tab:results-synthetic}).
Specifically, we study their vulnerability to the pitfalls proposed and assess their relative performance.
One by one, we visit each pitfall in the following sections.

\subsubsection{P1.1 (Missing) }

\QKCG, \QEnt and \QBALD show significant gains over competitors in the rare modality setting.
However, the benefits are not reliable in the case of \QKCG.
\QEnt performs best overall, taking the second place in the mid-label and the first place in the high-label regime.

\subsubsection{P2 \& P3.1 (Share)}

No method reliably improves the weak modality performance in the share setting.
\QRand performs as well as any specialized strategy.

\subsubsection{P3.2 (Unique)}

With respect to the partial information modality, differences emerge in the low- and mid-label settings, where \QGRACE and \QKCG perform best.
In the high label regime, the scores of all methods converge to similar values with \QRand leading the comparison.

\subsubsection{P3.3 (Synergy)}

Query methods perform similar with \QKCG achieving the first place in the low- and mid-label regimes.
\QBADGE leads in the high-label setting, suggesting that the relative advantage of different strategies depends on the label budget.
No method consistently and substantially outperforms its competitors.

\subsubsection{Summary}
\begin{table}[tb]
\caption{Summary of Query Method Rankings on Synthetic Data. In order to compare the query methods, we sum the ranks obtained by the query methods for all datasets. The sum of ranks indicates the overall performance. Lower is better. The best entries are highlighted in \textbf{bold}.}\label{tab:ranking-synthetic-summary}

\centering
\begin{tabularx}{\textwidth}{Xccccccc}
\toprule
Label Regime & \multicolumn{7}{c}{Query Method} \\
& \QBADGE & \QBALD & \QBMMAL & \QEnt & \QGRACE & \QKCG & \QRand \\\midrule
Low                         & 16    & 22   & 14    & 17      & 15    & \textbf{11}  & 17     \\
Mid                         & 22    & 21   & 14    & 17      & 17    & \textbf{6}   & 15     \\
High                        & 14    & \textbf{11}   & 15    & 21      & 15    & 20  & 16     \\\bottomrule
\end{tabularx}
\end{table}

The preceding analysis shows that the imbalances introduced by our pitfalls (\cref{sec:pitfalls-effect}) are largely unaddressed by existing query methods, with the exception of P1.1.
Furthermore, no single query method consistently outperforms the others across all settings and label regimes. 
To provide an aggregate view, we report summed ranks across weak-modality evaluations in \cref{tab:ranking-synthetic-summary}. 
Overall, \QKCG achieves the best aggregate rank in the low- and mid-label regimes, but its advantage disappears in the high-label regime, where \QBALD ranks first. 
This instability further underscores the absence of a reliable strategy for multimodal AL.

\begin{table}[htb]
\caption{Query Method Comparison on Real-World Data. \AUC per query method on real-world datasets across all label regimes. Results are reported separately for each modality combination. Best entries are highlighted in \textbf{bold}, the weak modality in \textit{italics}.}\label{tab:results-rw}
\centering
\begin{tabularx}{\textwidth}{llXccccccc}
\toprule
\multicolumn{2}{c}{\multirow{2}{1.5cm}{\textbf{Dataset Regime}}} & \textbf{Metric} & \multicolumn{7}{c}{\textbf{AULC per Query Method}}\\\cmidrule{4-10}
                         & & & BADGE     & BALD      & BMMAL     & Entropy   & GRACE     & KCG       & Random    \\
\midrule

\multirow{9}{*}{\rotatebox{90}{Food101}} & \multirow{3}{*}{\rotatebox{90}{Low}}
 & \textit{Image}   & 0.98  & 0.90  & 0.94  & 0.83  & 0.83  & 0.88    & 0.97  \\
&& Both             & 2.42  & 3.45  & 3.38  & 3.42  & 3.44  & 3.05    & 2.54  \\
&& Text             & 1.80  & 2.96  & 2.86  & 3.01  & 3.01  & 2.50    & 1.90  \\\cmidrule{3-10}

& \multirow{3}{*}{\rotatebox{90}{Mid}}
 & \textit{Image}   & 0.96  & 0.92  & 0.93  & 0.78  & 0.94  & 0.93    & 1.04  \\
&& Both             & 3.42  & 4.47  & 3.22  & 3.70  & 3.14  & 3.43    & 3.23  \\
&& Text             & 2.88  & 4.12  & 2.67  & 3.30  & 2.57  & 2.90    & 2.68  \\\cmidrule{3-10}

& \multirow{3}{*}{\rotatebox{90}{High}}
 & \textit{Image}   & 0.57  & 0.50  & 0.62  & 0.56  & 0.68  & 0.67    & 0.60 \\
&& Both             & 2.10  & 2.16  & 2.72  & 2.47  & 1.72  & 2.41    & 2.45 \\
&& Text             & 1.88  & 1.99  & 2.55  & 2.29  & 1.23  & 2.17    & 2.23 \\\midrule


\multirow{9}{*}{\rotatebox{90}{MIMIC}} & \multirow{3}{*}{\rotatebox{90}{Low}}
 & \textit{ECG}         & 4.60  & 4.61  & 4.62  & 4.60  & 4.63  & 4.64    & \textbf{4.73}  \\
&& Both                 & 6.51  & \textbf{6.52}  & 6.51  & 6.49  & 6.43  & 6.42    & 6.51  \\
&& Bloodlabs            & \textbf{6.55}  & \textbf{6.55}  & 6.52  & 6.53  & 6.44  & 6.03    & 6.54  \\\cmidrule{3-10}

& \multirow{3}{*}{\rotatebox{90}{Mid}}
 & \textit{ECG}         & 4.71  & 4.87  & 4.74  & 4.67  & 4.82  & 4.62    & \textbf{5.03}  \\
&& Both                 & 6.55  & \textbf{6.59}  & 6.55  & 6.57  & 6.45  & 6.49    & 6.58  \\
&& Bloodlabs            & 6.54  & \textbf{6.61}  & 6.54  & 6.56  & 6.40  & 6.27    & \textbf{6.61}  \\\cmidrule{3-10}

& \multirow{3}{*}{\rotatebox{90}{High}}
 & \textit{ECG}         & 4.85  & 5.13  & 4.87  & 4.74  & 4.76  & 4.76    & \textbf{5.23} \\
&& Both                 & 6.62  & 6.66  & \textbf{6.67}  & 6.61  & 6.56  & 6.61    & \textbf{6.67} \\
&& Bloodlabs            & 6.59  & 6.65  & 6.64  & 6.59  & 6.53  & 6.52    & \textbf{6.66} \\\midrule

\bottomrule
\end{tabularx}
\end{table}

\subsection{Query Method Comparison on Real-World Data}\label{sec:results-rw}

We now validate whether the patterns observed on synthetic data carry over to real-world multimodal datasets.

\subsubsection{Food101}

The Food101 dataset shows imbalance problems similar to those seen in the share setting: the image modality is considerably outperformed and the performance of the joint evaluation tracks the stronger modality.
Among query methods, \QBALD performs well across regimes, achieving the highest joint \AUC in the low- and mid-label settings. 
However, in the high-label regime, \QBMMAL takes the lead, while \QGRACE, a multimodal method, yields the lowest joint score. 
No method reliably improves the weak (image) modality.

\subsubsection{MIMIC}

On MIMIC-IV, Bloodlabs constitute the stronger modality, substantially outperforming ECGs across all regimes. 
The joint \AUC closely follows the Bloodlabs-only performance, again confirming the modality imbalance pattern. 
Notably, \QRand is competitive with or outperforms all query methods in most settings, echoing findings from the synthetic experiments.

\subsubsection{Summary}
Overall, the real-world results corroborate our synthetic findings: modality imbalance persists, existing query methods do not mitigate it, and multimodal strategies offer no consistent advantage over unimodal baselines.

\subsubsection{Limitations}
Our framework focuses on the two-modality setting, and while the identified pitfalls generalize conceptually, their interactions may grow more complex with additional modalities. Furthermore, all experiments use a late-fusion architecture; different fusion strategies, such as early fusion or cross-attention, may interact differently with the pitfalls and alter the relative performance of query methods.
Extending this benchmark to richer architectures and more than two modalities is a natural direction for future work.

%% file: sections/80_discussion.tex

%% file: sections/90_conclusion.tex
\section{Conclusion}

In this work, we introduced a controlled benchmark for evaluating AL methods in multimodal settings. 
Through carefully designed synthetic datasets, our framework isolates three key pitfalls: missing modalities, modality imbalance, and varying modality interaction structures. 
Our work is complemented by validation of our findings on real-world data.

Our experiments reveal a consistent pattern: multimodal models predominantly rely on a single modality, developing imbalanced representations that degrade performance when the weaker modality carries task-relevant information. 
This effect emerges across all pitfall settings and is not mitigated by any of the query strategies we evaluated. 
While some methods, notably \QKCG and \QBALD, perform comparatively well in aggregate, their advantages are inconsistent across settings. 
Multimodal query strategies do not provide clear advantages over unimodal ones in our experiments.

These findings point to a fundamental limitation: current AL query strategies are modality-agnostic in practice, even when designed with multimodal data in mind. 
We believe that effective multimodal AL will require query methods that explicitly monitor and correct modality imbalance during the acquisition process. 
The proposed benchmark provides a controlled foundation for developing and evaluating such methods.

\begin{credits}
\subsubsection{\ackname} We thank Thomas Wolf (Bayer AG) for initiating this collaborative work between Bayer AG and German Cancer Research Center (DKFZ) as well as his valuable input. This work was partially funded by the Bayer AG Life Science Collaboration Project `Embrace Uncertainty'.

\end{credits}

%% file: sections/93_app_params.tex
\section{Experimental Details}\label{sec:settings}

\subsection{MIMIC Dataset}\label{sec:settings-mimic}
To create the dataset, we collect all admissions and gather bloodlabs and ECGs that are taken between 24h prior and 24h post admission time. 
Since some patients are admitted to the hospital times, these patients will be featured in multiple samples.
We acknowledge this in our training and test splits, by randomly choosing 30,000 subjects for testing.

Bloodlabs contain measurements of various substances of the blood.
Since not all substances are measured at every time, we follow \cite{saportaContrastingSymileSimple2024} and choose only the 50 most frequent measurements. 
These are: Hematocrit, Platelet Count, Creatinine, Potassium, Hemoglobin, White Blood Cells, MCHC, Red Blood Cells, MCV, MCH, RDW, Urea Nitrogen, Sodium, Chloride, Bicarbonate, Anion Gap, Glucose, Magnesium, Calcium Total, Phosphate, INR(PT), PT, PTT, Basophils, Neutrophils, Monocytes, Eosinophils, Lymphocytes, RDW-SD, H, L, I, Alanine Aminotransferase (ALT), Asparate Aminotransferase (AST), Lactate, Alkaline Phosphatase, Bilirubin Total, pH, Albumin, Base Excess, pO2, Calculated Total CO2, pCO2, Absolute Neutrophil Count, Absolute Eosinophil Count, Absolute Monocyte Count, Absolute Basophil Count, Absolute Lymphocyte Count, Creatine Kinase (CK), Immature Granulocytes.
Even within those, not all substances are measured at all times.
We employ quantile scaling and median imputation for bloodlabs.

For ECGs, we filter out empty ECGs and those containing NaN values. 
We rescale the ECGs to have values in [-1, 1].

\subsection{Model Architecture}\label{sec:settings-arch}

For our experiments, we use a simple architecture with uni-modal encoders. All uni-modal latents are fed through a non-linear projection layer to create feature vectors of equal size. The resulting feature vectors are then fused by a transformer encoder and passed to a linear classifier.

For the image modalities, we use the EfficientNetB0 \cite{tanEfficientNetRethinkingModel2020} architecture. In the share experiment, we adopt the first convolutional layer to accept one-channel inputs. For the textual inputs of the Food101 dataset, we use the MiniLM \cite{wangMiniLMDeepSelfAttention2020} architecture\footnote{\url{https://huggingface.co/sentence-transformers/all-MiniLM-L6-v2}}.  For MIMIC-IV, bloodlabs are fed through a two-layer MLP. ECGs are processed using a one-dimensional version of the EfficientNetB0. In MIMIC, we additionally employ dropout in the projection layers.

\subsection{Training Settings}\label{sec:settings-training}
\begin{table}[]
\centering
\caption{Training Settings}\label{tab:params-training}

This table shows the results of the hyperparameter search described in \cref{sec:settings-training}. An asterisk (*) denotes that the listed settings are applied to all label regimes.

\begin{tabularx}{\textwidth}{XXXXXXX}
\toprule
Dataset                & \begin{tabular}[c]{@{}l@{}}Label\\ Regime\end{tabular} & Epochs              & \begin{tabular}[c]{@{}l@{}}Batch\\ Size\end{tabular} & \begin{tabular}[c]{@{}l@{}}Learning\\ Rate\end{tabular} & \begin{tabular}[c]{@{}l@{}}Weight\\ Decay\end{tabular} & Augment.        \\\midrule \\ 

Missing                & *                                                      & 60                  & 128                                                  & 0.1                                                     & 5e-4                                                   & RandAug            \\\midrule\\ 

Share                  & *                                                      & 200                 & 1024                                                 & 0.1                                                     & 5e-3                                                   & RandAug            \\\midrule\\ 

Unique                 & *                                                      & 60                  & 128                                                  & 0.1                                                     & 5e-4                                                   & Basic     \\\midrule\\ 

Synergy                & *                                                      & 60                  & 128                                                  & 0.1                                                     & 5e-4                                                   & Basic     \\\midrule\\ 

Food101                & *                                                      & 60                  & 128                                                  & 0.1                                                     & 5e-4                                                   & RandAug \\\midrule\\ 

\multirow{3}{*}{MIMIC} & Low                                                    & \multirow{3}{*}{60} & \multirow{3}{*}{256}                                 & 0.01                                                    & 5e-5                                                   & None                 \\\\
                       & Mid                                                    &                     &                                                      & 0.01                                                   & 5e-4                                                   & None                 \\\\
                       & High                                                   &                     &                                                      & 0.001                                                   & 5e-5                                                   & None                 \\\\ \bottomrule
\end{tabularx}
\end{table}

In this section, we describe in detail how the hyperparameters for model training were selected.
The number of epochs, batch size and the optimizer and learning rate scheduler are pre-selected for each experiment.
For all experiments, we use stochastic gradient descent with a cosine annealing learning rate scheduler and ten warm-up epochs.

The learning rate, weight decay and augmentations are chosen via grid search.
Possible values are $[0.1, 0.01, 0.001]$ for the learning rate, $[5e-3, 5e-4]$ for weight decay and [Basic, RandAugMC] for the augmentations.
Basic is a combination of random horizontal flipping and random crops.
RandAug additionally applies various strong augmentations following \cite{cubukRandaugmentPracticalAutomated2020}.

Augmentations are only applied to the image modalities.
If a dataset has two image modalities as inputs, the augmentations are used for both modalities.
Other types of modalities are not augmented.

\Cref{tab:params-training} shows the training settings used in our experiments. 
For MIMIC, we additionally use class-balanced sampling.

\subsection{Active Learning Settings}\label{sec:settings-al}

\begin{table}[]
\centering
\caption{Active Learning Settings}\label{tab:params-al}
\begin{tabularx}{\textwidth}{XXXXXX}
\toprule
Dataset                                   & \begin{tabular}[c]{@{}l@{}}Label\\ Regime\end{tabular} & \begin{tabular}[c]{@{}l@{}}Initial\\ Budget\end{tabular} & \begin{tabular}[c]{@{}l@{}}Validation\\ Size\end{tabular} & \begin{tabular}[c]{@{}l@{}}Acquisition\\ Size\end{tabular} & Iterations \\\midrule\\
\multirow[t]{3}{1.5cm}{Missing Share}     & Low          & 50             & 250                 & 50               & 10                   \\\\
                                          & Mid          & 250            & 1250                & 250              & 10                   \\\\
                                          & High         & 1000           & 5000                & 1000             & 10                   \\\midrule\\ 
\multirow[t]{3}{1.5cm}{Unique Synergy MIMIC} & Low          & 500            & 2500                & 500              & 10                   \\\\
                                          & Mid          & 1000           & 5000                & 1000             & 10                   \\\\
                                          & High         & 5000           & 5000                & 5000             & 10                   \\\midrule\\ 
\multirow[t]{3}{1.5cm}{Food101}           & Low          & 505            & 2525                & 505              & 10                   \\\\
                                          & Mid          & 1010           & 5050                & 1010             & 10                   \\\\
                                          & High         & 5050           & 5050                & 5050             & 5                    \\\bottomrule
\end{tabularx}
\end{table}

Following \cite{luthNavigatingPitfallsActive2023}, we investigate different label-regimes in order to verify that the query methods work under a variety of conditions.
These label regimes change encompass different starting budgets, validation set sizes, acquisition sizes and numbers of iterations.
We follow their rule of thumb to select the starting budget and acquisition size based on $5*C$ for the 'low-label', $25*C$ for the 'mid-label' and $100*C$ for the 'high-label' regimes, where $C$ denotes the number of classes.
For \emph{MIMIC}, we make an exception to this rule.
\emph{MIMIC} is a very-large dataset and the task is binary-classification.
Thus, using this rule of thumb would lead to very small budgets and query sizes in comparison to the dataset size.
Therefore, we adopt the values from the \emph{Unique} and \emph{Synergy} settings.

The number of iterations is set to ten.
The only exception is the 'high-label' regime of the \emph{Food101} dataset, where we only conduct five iterations to stay below half of the full dataset size in the final iteration.

%% file: sections/94_app_eval.tex
\clearpage
\section{Detailed Results}\label{sec:detailed-results}

\newcommand{\resultsTableCaption}[1]{Shows the \AUC ± standard deviation of all query methods on the #1 dataset. Higher is better.}
\newcommand{\resultsFigureCaption}[1]{
Shows the test results of all query methods on the #1 dataset. 
Top: shows the balanced test accuracy over the course of active learning experiments. 
The standard deviation is drawn using a lighter shade. 
Bottom: summarizes the plots above using the \AUC. Standard deviation is added to the bars. For each query method and modality, there are three bars representing the low-, mid- and high-label regimes from left to right.
In both subfigures, higher values represent a better performance.
}

\subsection{Missing}
\begin{figure}
    {
        \centering
        \begin{subfigure}{\textwidth}
            \includegraphics[width=\textwidth]{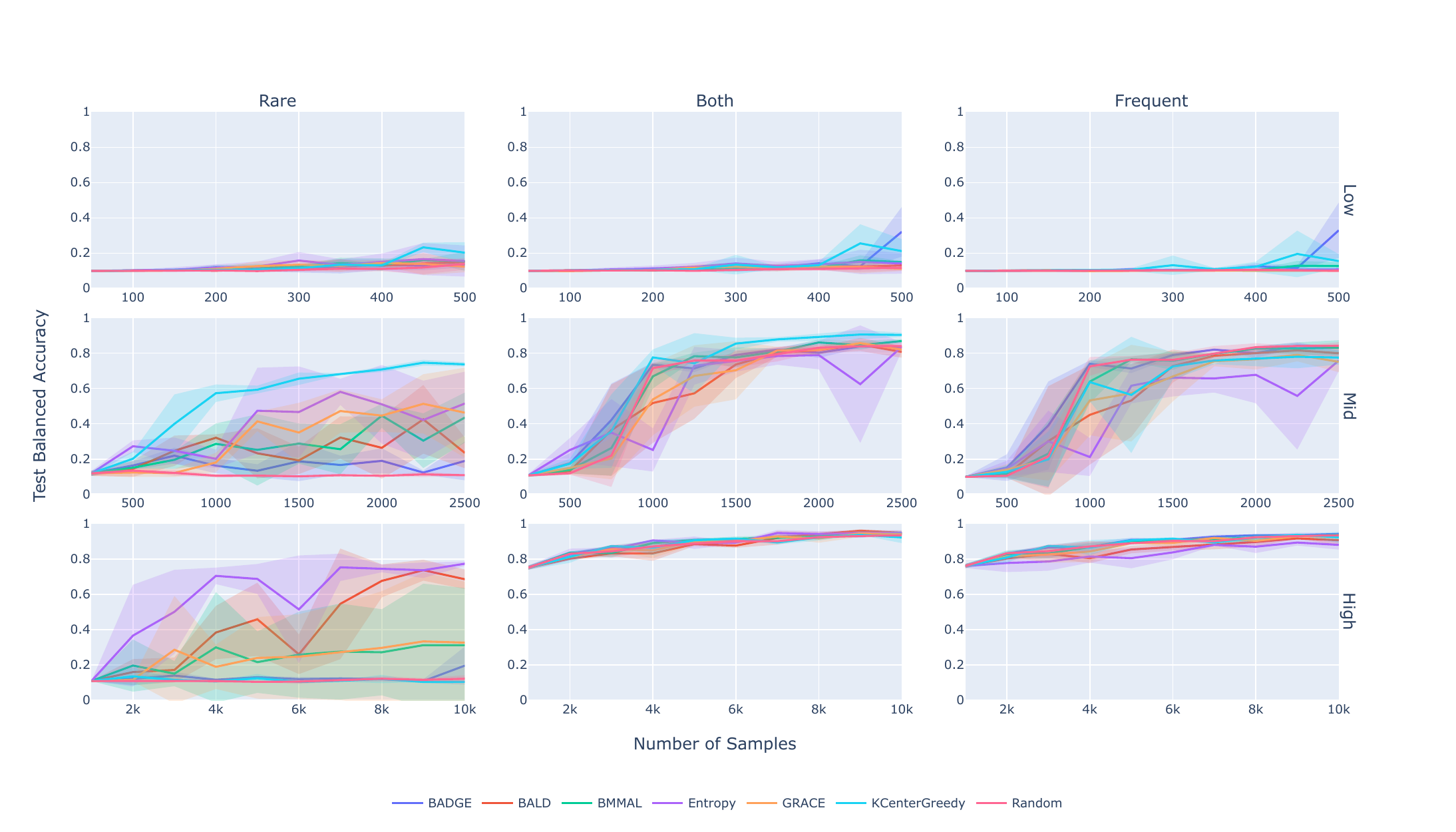}
        \end{subfigure}
        \begin{subfigure}{\textwidth}
            \includegraphics[width=\textwidth]{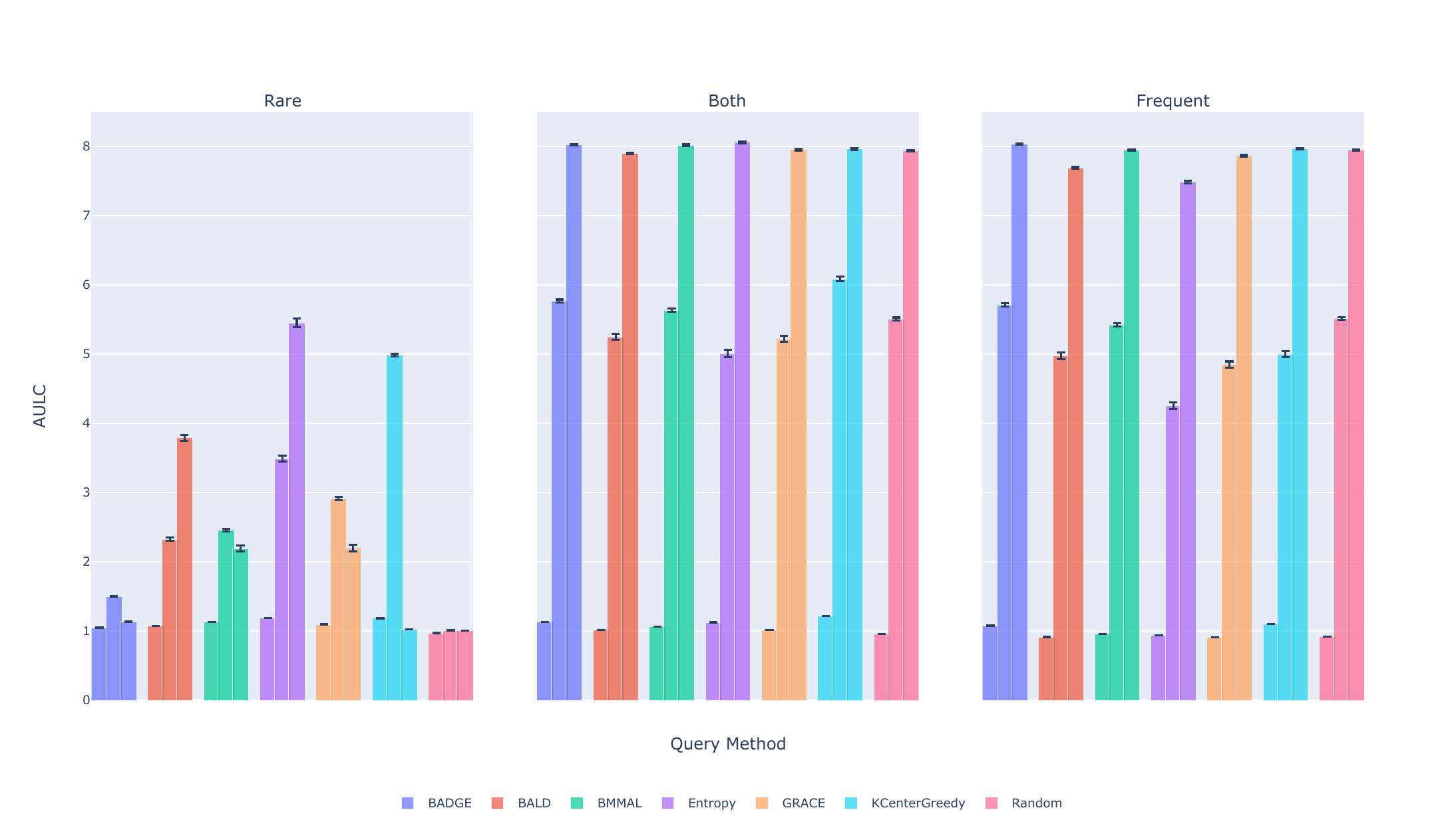}
        \end{subfigure}
    }
    \caption{Results for Missing with ModDrop}\label{fig:results-missing-dm}
    \resultsFigureCaption{Missing}
\end{figure}

\begin{table}
\caption{Results for Missing with ModDrop}\label{tab:results-missing-md}
\resultsTableCaption{Missing}
\centering
\begin{tabularx}{\textwidth}{XXXXX}
\toprule
\multicolumn{1}{l}{\textbf{Regime}} & \multicolumn{1}{l}{\textbf{QM}} & \multicolumn{1}{l}{\textbf{Rare}} & \multicolumn{1}{l}{\textbf{Both}} & \multicolumn{1}{l}{\textbf{Frequent}} \\\midrule\\
Low & BADGE & 1.05 ± 0.16 & 1.13 ± 0.17 & 1.08 ± 0.13 \\
 & BALD & 1.07 ± 0.10 & 1.02 ± 0.06 & 0.91 ± 0.02 \\
 & BMMAL & 1.13 ± 0.15 & 1.06 ± 0.09 & 0.96 ± 0.06 \\
 & Entropy & 1.19 ± 0.33 & 1.12 ± 0.24 & 0.94 ± 0.05 \\
 & GRACE & 1.10 ± 0.19 & 1.02 ± 0.13 & 0.91 ± 0.02 \\
 & KCenterGreedy & 1.19 ± 0.14 & 1.22 ± 0.24 & 1.10 ± 0.26 \\
 & Random & 0.97 ± 0.11 & 0.96 ± 0.06 & 0.92 ± 0.03 \\ \\\midrule \\
Mid & BADGE & 1.50 ± 0.46 & 5.77 ± 0.36 & 5.71 ± 0.43 \\
 & BALD & 2.33 ± 0.91 & 5.25 ± 0.82 & 4.98 ± 0.96 \\
 & BMMAL & 2.46 ± 0.93 & 5.63 ± 0.39 & 5.43 ± 0.47 \\
 & Entropy & 3.49 ± 1.28 & 5.01 ± 1.03 & 4.26 ± 1.11 \\
 & GRACE & 2.91 ± 0.84 & 5.22 ± 0.77 & 4.85 ± 0.92 \\
 & KCenterGreedy & 4.99 ± 0.36 & 6.09 ± 0.56 & 5.00 ± 0.86 \\
 & Random & 1.01 ± 0.07 & 5.51 ± 0.36 & 5.52 ± 0.38 \\ \\\midrule \\
High & BADGE & 1.13 ± 0.13 & 8.02 ± 0.10 & 8.03 ± 0.10 \\
 & BALD & 3.79 ± 1.11 & 7.90 ± 0.16 & 7.69 ± 0.18 \\
 & BMMAL & 2.19 ± 1.98 & 8.02 ± 0.12 & 7.95 ± 0.13 \\
 & Entropy & 5.45 ± 1.11 & 8.06 ± 0.15 & 7.49 ± 0.31 \\
 & GRACE & 2.20 ± 2.07 & 7.96 ± 0.15 & 7.86 ± 0.17 \\
 & KCenterGreedy & 1.03 ± 0.12 & 7.96 ± 0.15 & 7.97 ± 0.15 \\
 & Random & 1.01 ± 0.08 & 7.94 ± 0.10 & 7.95 ± 0.11 \\ \\\bottomrule \\
\end{tabularx}
\end{table}

\begin{figure}
    {
        \centering
        \begin{subfigure}{\textwidth}
            \includegraphics[width=\textwidth]{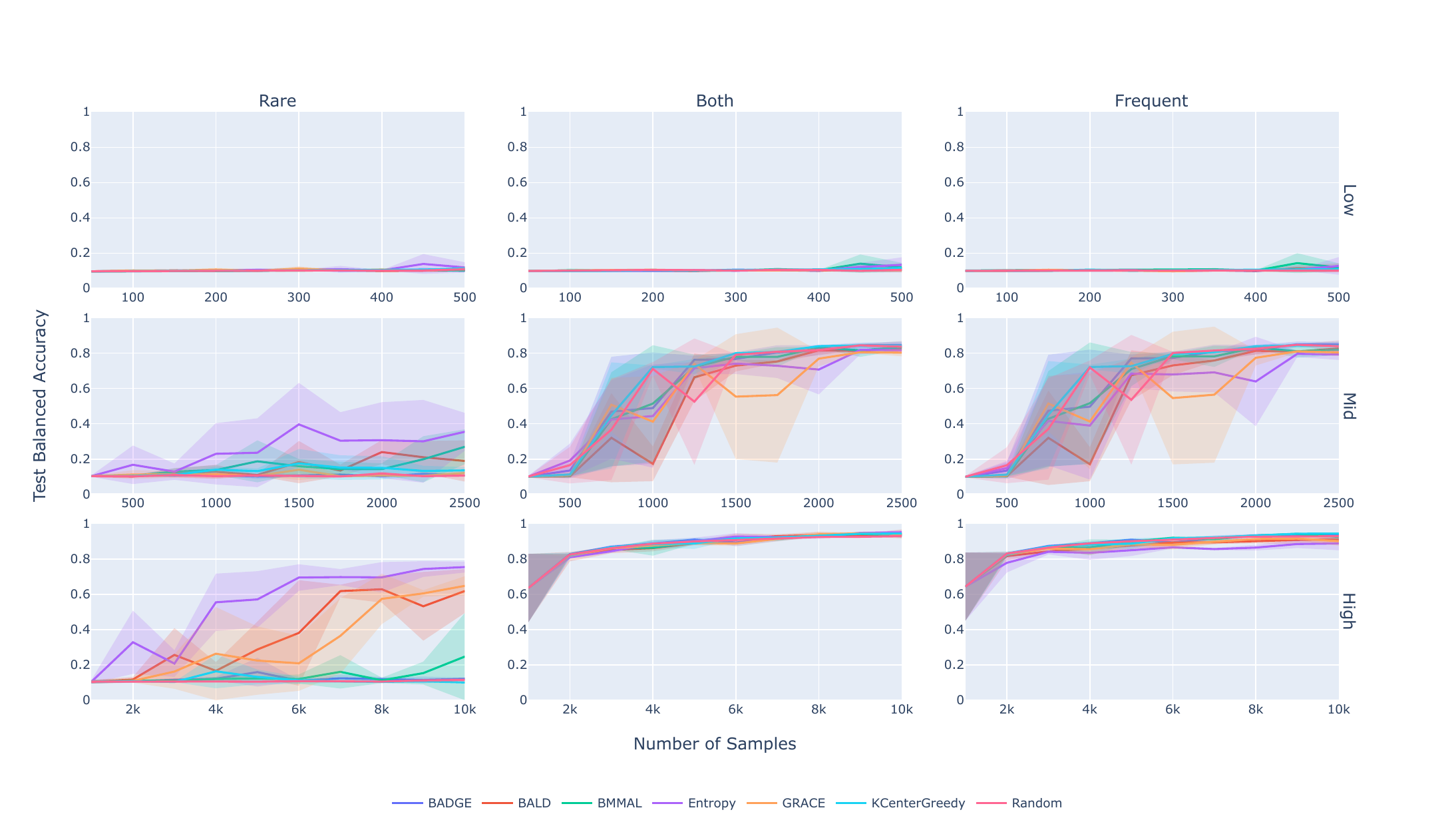}
        \end{subfigure}
        \begin{subfigure}{\textwidth}
            \includegraphics[width=\textwidth]{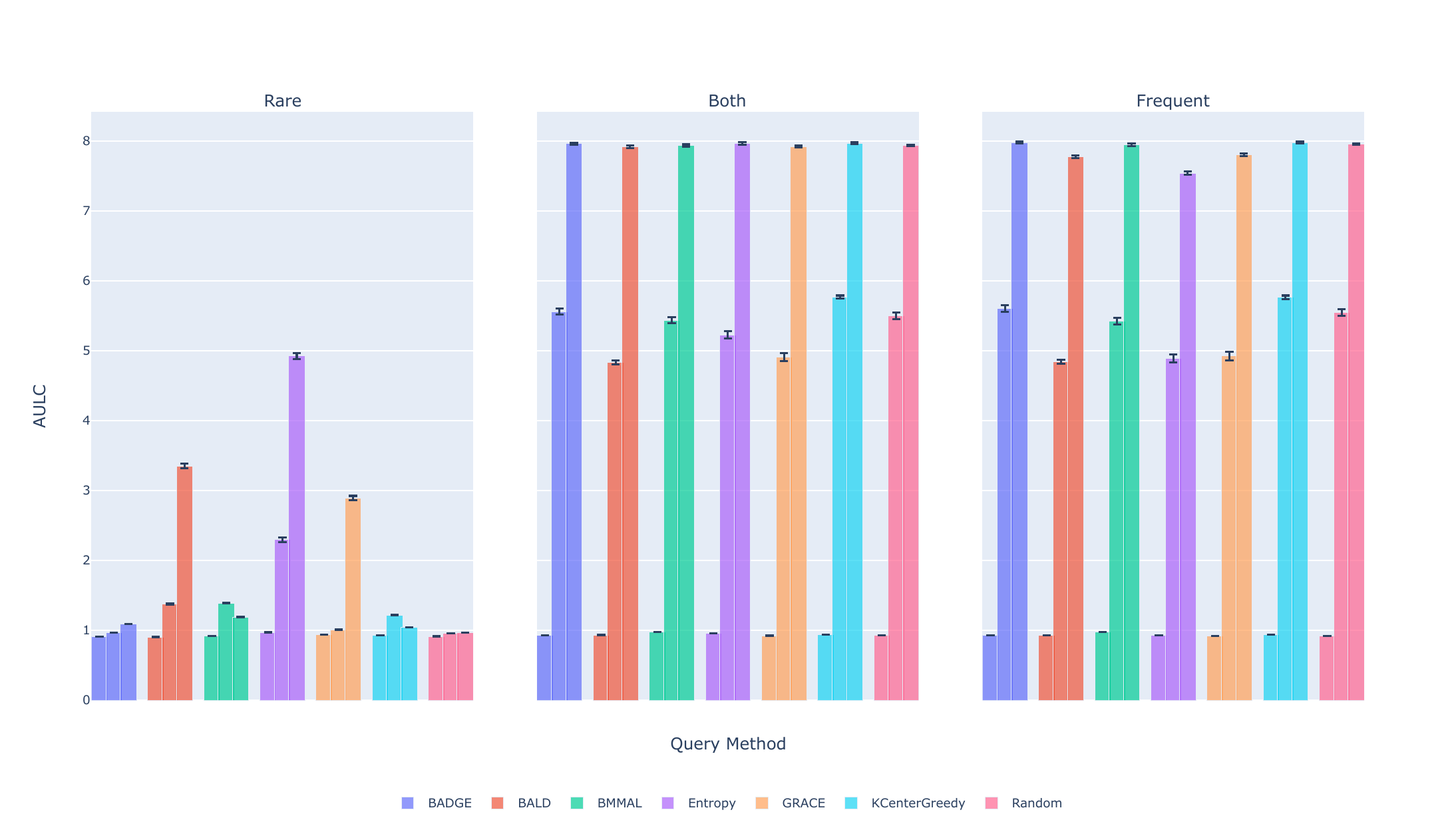}
        \end{subfigure}
    }
    \caption{Results for Missing without ModDrop}\label{fig:results-missing}
    \resultsFigureCaption{Missing}
\end{figure}

\begin{table}
\caption{Results for Missing without ModDrop}\label{tab:results-missing}
\resultsTableCaption{Missing}
\centering
\begin{tabularx}{\textwidth}{XXXXX}
\toprule
\multicolumn{1}{l}{\textbf{Regime}} & \multicolumn{1}{l}{\textbf{QM}} & \multicolumn{1}{l}{\textbf{A}} & \multicolumn{1}{l}{\textbf{Both}} & \multicolumn{1}{l}{\textbf{B}} \\\midrule\\
Low & BADGE & 0.91 ± 0.04 & 0.93 ± 0.05 & 0.93 ± 0.05 \\
 & BALD & 0.91 ± 0.03 & 0.93 ± 0.05 & 0.93 ± 0.05 \\
 & BMMAL & 0.92 ± 0.02 & 0.98 ± 0.10 & 0.98 ± 0.10 \\
 & Entropy & 0.97 ± 0.11 & 0.96 ± 0.07 & 0.93 ± 0.05 \\
 & GRACE & 0.94 ± 0.05 & 0.92 ± 0.03 & 0.92 ± 0.02 \\
 & KCenterGreedy & 0.93 ± 0.03 & 0.94 ± 0.03 & 0.94 ± 0.03 \\
 & Random & 0.92 ± 0.03 & 0.93 ± 0.04 & 0.92 ± 0.05 \\ \\\midrule \\
Mid & BADGE & 0.97 ± 0.05 & 5.56 ± 0.83 & 5.60 ± 0.84 \\
 & BALD & 1.38 ± 0.45 & 4.84 ± 0.57 & 4.85 ± 0.60 \\
 & BMMAL & 1.39 ± 0.47 & 5.44 ± 0.79 & 5.42 ± 0.85 \\
 & Entropy & 2.30 ± 1.43 & 5.23 ± 0.95 & 4.89 ± 1.18 \\
 & GRACE & 1.01 ± 0.15 & 4.91 ± 1.24 & 4.93 ± 1.29 \\
 & KCenterGreedy & 1.22 ± 0.31 & 5.77 ± 0.40 & 5.77 ± 0.42 \\
 & Random & 0.96 ± 0.04 & 5.50 ± 0.84 & 5.55 ± 0.85 \\ \\\midrule \\
High & BADGE & 1.09 ± 0.17 & 7.97 ± 0.17 & 7.98 ± 0.17 \\
 & BALD & 3.35 ± 1.05 & 7.93 ± 0.25 & 7.78 ± 0.24 \\
 & BMMAL & 1.19 ± 0.36 & 7.94 ± 0.20 & 7.95 ± 0.21 \\
 & Entropy & 4.93 ± 0.85 & 7.97 ± 0.20 & 7.55 ± 0.31 \\
 & GRACE & 2.89 ± 1.13 & 7.92 ± 0.18 & 7.81 ± 0.22 \\
 & KCenterGreedy & 1.04 ± 0.20 & 7.97 ± 0.19 & 7.98 ± 0.19 \\
 & Random & 0.97 ± 0.05 & 7.94 ± 0.13 & 7.96 ± 0.13 \\ \\\bottomrule \\
\end{tabularx}
\end{table}

\clearpage

\subsection{Share}
\begin{figure}
    {
        \centering
        \begin{subfigure}{\textwidth}
            \includegraphics[width=\textwidth]{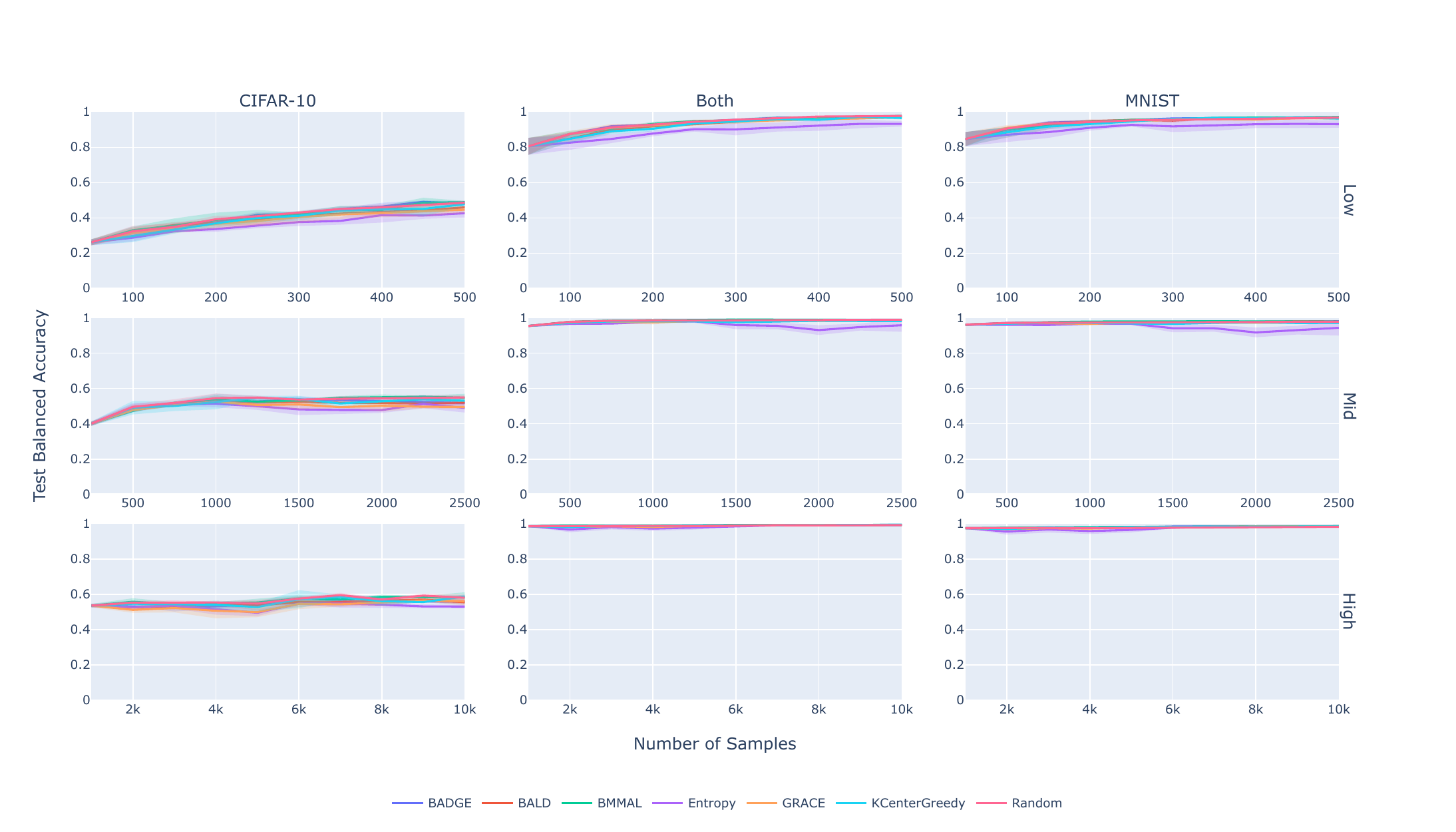}
        \end{subfigure}
        \begin{subfigure}{\textwidth}
            \includegraphics[width=\textwidth]{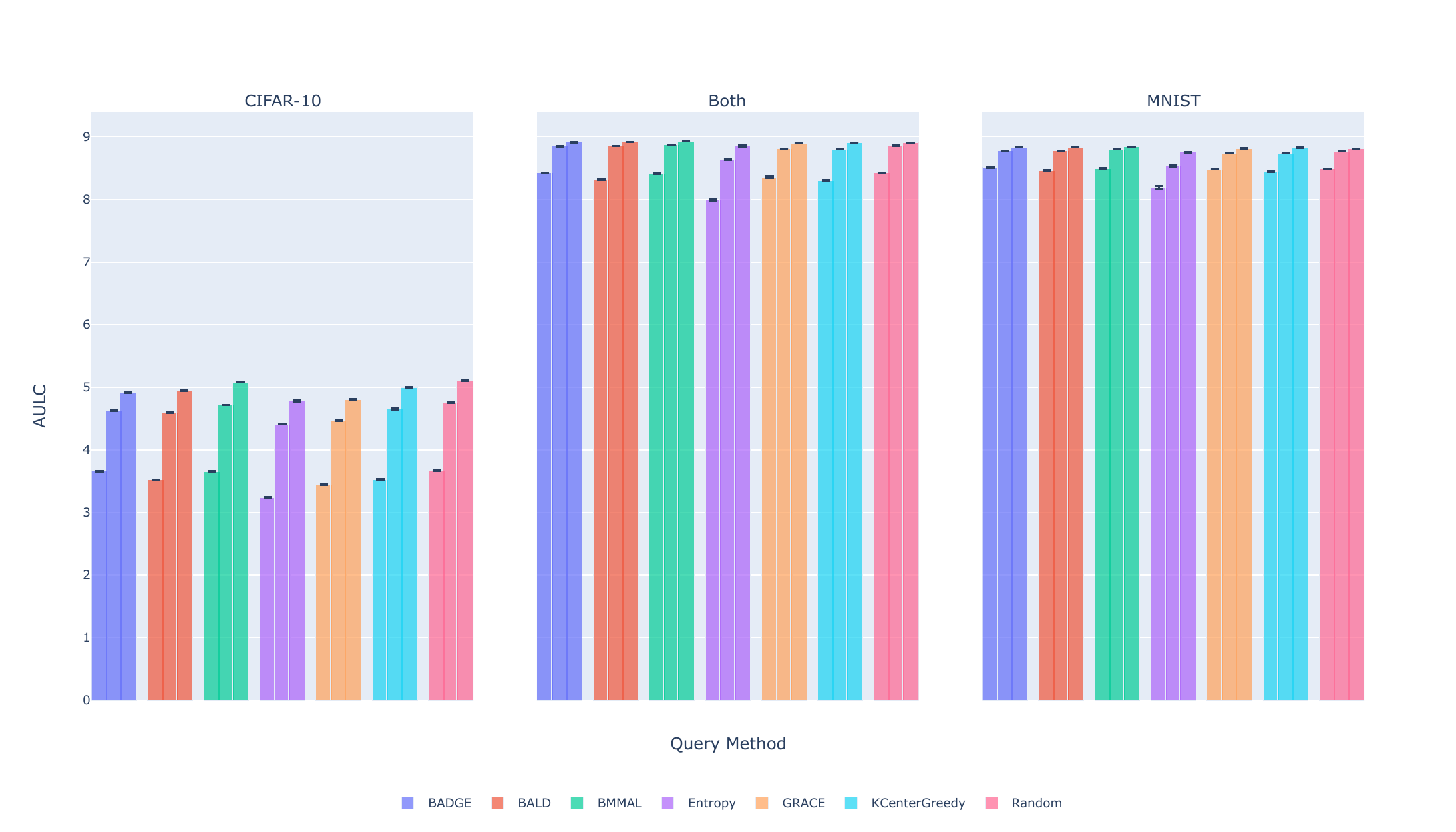}
        \end{subfigure}
    }
    \caption{Results for Share with ModDrop}\label{fig:results-share-md}
    \resultsFigureCaption{Share}
\end{figure}

\begin{table}
\caption{Results for Share with ModDrop}\label{tab:results-share-md}
\resultsTableCaption{Share}
\centering
\begin{tabularx}{\textwidth}{XXXXX}
\toprule
\multicolumn{1}{l}{\textbf{Regime}} & \multicolumn{1}{l}{\textbf{QM}} & \multicolumn{1}{l}{\textbf{CIFAR-10}} & \multicolumn{1}{l}{\textbf{Both}} & \multicolumn{1}{l}{\textbf{MNIST}} \\\midrule\\
Low & BADGE & 3.66 ± 0.12 & 8.42 ± 0.06 & 8.51 ± 0.07 \\
 & BALD & 3.52 ± 0.16 & 8.32 ± 0.11 & 8.46 ± 0.12 \\
 & BMMAL & 3.66 ± 0.24 & 8.41 ± 0.09 & 8.49 ± 0.07 \\
 & Entropy & 3.24 ± 0.18 & 7.99 ± 0.24 & 8.19 ± 0.24 \\
 & GRACE & 3.45 ± 0.23 & 8.36 ± 0.13 & 8.48 ± 0.09 \\
 & KCenterGreedy & 3.53 ± 0.22 & 8.30 ± 0.13 & 8.45 ± 0.12 \\
 & Random & 3.66 ± 0.14 & 8.42 ± 0.08 & 8.49 ± 0.06 \\ \\\midrule \\
Mid & BADGE & 4.62 ± 0.10 & 8.85 ± 0.03 & 8.78 ± 0.03 \\
 & BALD & 4.59 ± 0.10 & 8.86 ± 0.02 & 8.78 ± 0.03 \\
 & BMMAL & 4.72 ± 0.10 & 8.87 ± 0.02 & 8.80 ± 0.02 \\
 & Entropy & 4.41 ± 0.19 & 8.64 ± 0.14 & 8.54 ± 0.14 \\
 & GRACE & 4.47 ± 0.13 & 8.81 ± 0.05 & 8.74 ± 0.05 \\
 & KCenterGreedy & 4.65 ± 0.20 & 8.80 ± 0.03 & 8.73 ± 0.03 \\
 & Random & 4.76 ± 0.13 & 8.86 ± 0.02 & 8.77 ± 0.02 \\ \\\midrule \\
High & BADGE & 4.91 ± 0.12 & 8.91 ± 0.02 & 8.83 ± 0.03 \\
 & BALD & 4.94 ± 0.15 & 8.92 ± 0.01 & 8.84 ± 0.01 \\
 & BMMAL & 5.08 ± 0.12 & 8.93 ± 0.01 & 8.84 ± 0.02 \\
 & Entropy & 4.78 ± 0.15 & 8.85 ± 0.06 & 8.75 ± 0.08 \\
 & GRACE & 4.80 ± 0.20 & 8.90 ± 0.03 & 8.81 ± 0.03 \\
 & KCenterGreedy & 5.00 ± 0.18 & 8.91 ± 0.02 & 8.83 ± 0.02 \\
 & Random & 5.10 ± 0.11 & 8.91 ± 0.01 & 8.81 ± 0.02 \\ \\\bottomrule \\
\end{tabularx}
\end{table}

\begin{figure}
    {
        \centering
        \begin{subfigure}{\textwidth}
            \includegraphics[width=\textwidth]{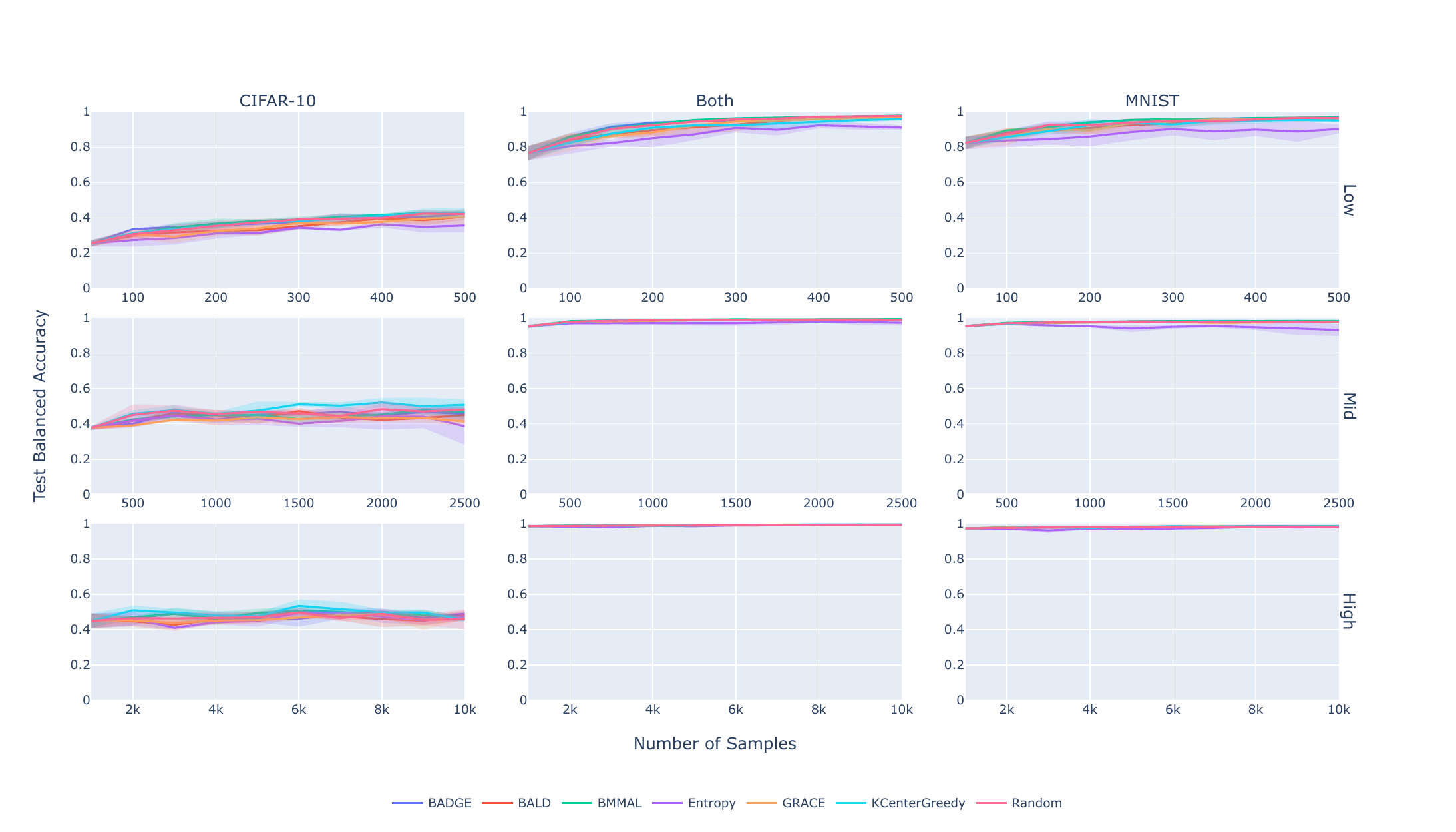}
        \end{subfigure}
        \begin{subfigure}{\textwidth}
            \includegraphics[width=\textwidth]{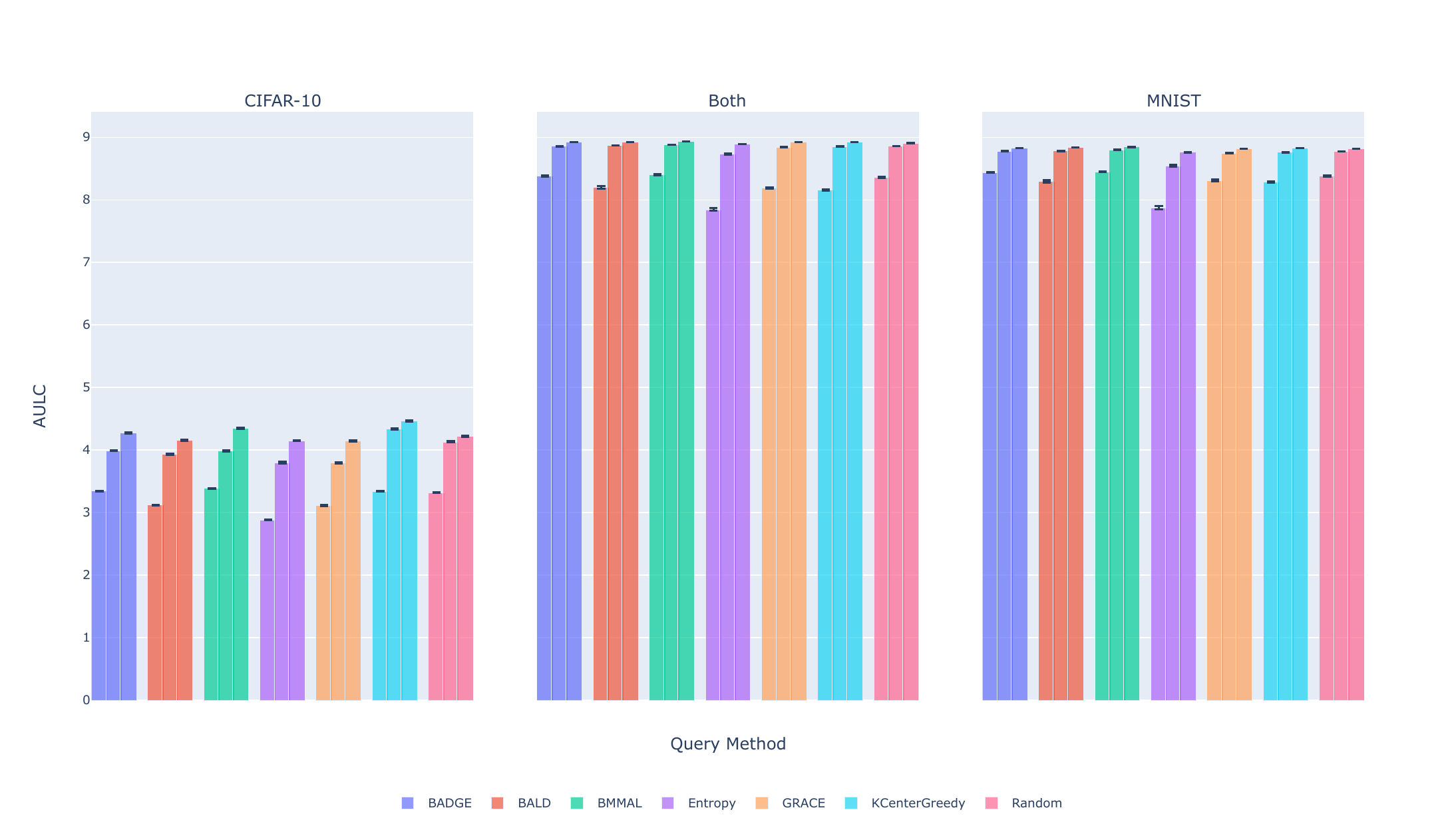}
        \end{subfigure}
    }
    \caption{Results for Share without ModDrop}\label{fig:results-share}
    \resultsFigureCaption{Share}
\end{figure}

\begin{table}
\caption{Results for Share without ModDrop}\label{tab:results-share}
\resultsTableCaption{Share}
\centering
\begin{tabularx}{\textwidth}{XXXXX}
\toprule
\multicolumn{1}{l}{\textbf{Regime}} & \multicolumn{1}{l}{\textbf{QM}} & \multicolumn{1}{l}{\textbf{CIFAR-10}} & \multicolumn{1}{l}{\textbf{Both}} & \multicolumn{1}{l}{\textbf{MNIST}} \\\midrule\\
Low & BADGE & 3.35 ± 0.13 & 8.38 ± 0.09 & 8.44 ± 0.09 \\
 & BALD & 3.12 ± 0.17 & 8.20 ± 0.22 & 8.30 ± 0.19 \\
 & BMMAL & 3.39 ± 0.14 & 8.40 ± 0.08 & 8.45 ± 0.07 \\
 & Entropy & 2.89 ± 0.21 & 7.84 ± 0.27 & 7.88 ± 0.38 \\
 & GRACE & 3.11 ± 0.22 & 8.19 ± 0.16 & 8.31 ± 0.16 \\
 & KCenterGreedy & 3.34 ± 0.13 & 8.16 ± 0.16 & 8.28 ± 0.17 \\
 & Random & 3.32 ± 0.18 & 8.36 ± 0.08 & 8.38 ± 0.09 \\ \\\midrule \\
Mid & BADGE & 3.99 ± 0.20 & 8.86 ± 0.02 & 8.78 ± 0.02 \\
 & BALD & 3.93 ± 0.16 & 8.87 ± 0.01 & 8.78 ± 0.02 \\
 & BMMAL & 3.99 ± 0.15 & 8.88 ± 0.01 & 8.80 ± 0.02 \\
 & Entropy & 3.80 ± 0.38 & 8.73 ± 0.10 & 8.54 ± 0.14 \\
 & GRACE & 3.79 ± 0.20 & 8.84 ± 0.03 & 8.75 ± 0.06 \\
 & KCenterGreedy & 4.34 ± 0.23 & 8.85 ± 0.03 & 8.76 ± 0.03 \\
 & Random & 4.13 ± 0.26 & 8.86 ± 0.01 & 8.77 ± 0.02 \\ \\\midrule \\
High & BADGE & 4.27 ± 0.23 & 8.93 ± 0.01 & 8.83 ± 0.02 \\
 & BALD & 4.16 ± 0.27 & 8.92 ± 0.01 & 8.84 ± 0.02 \\
 & BMMAL & 4.35 ± 0.21 & 8.93 ± 0.01 & 8.85 ± 0.02 \\
 & Entropy & 4.15 ± 0.22 & 8.89 ± 0.02 & 8.76 ± 0.06 \\
 & GRACE & 4.15 ± 0.20 & 8.92 ± 0.01 & 8.82 ± 0.03 \\
 & KCenterGreedy & 4.46 ± 0.23 & 8.92 ± 0.01 & 8.83 ± 0.02 \\
 & Random & 4.23 ± 0.24 & 8.91 ± 0.01 & 8.82 ± 0.02 \\\\\bottomrule\\
\end{tabularx}
\end{table}

\clearpage
\subsection{Unique}
\begin{figure}
    {
        \centering
        \begin{subfigure}{\textwidth}
            \includegraphics[width=\textwidth]{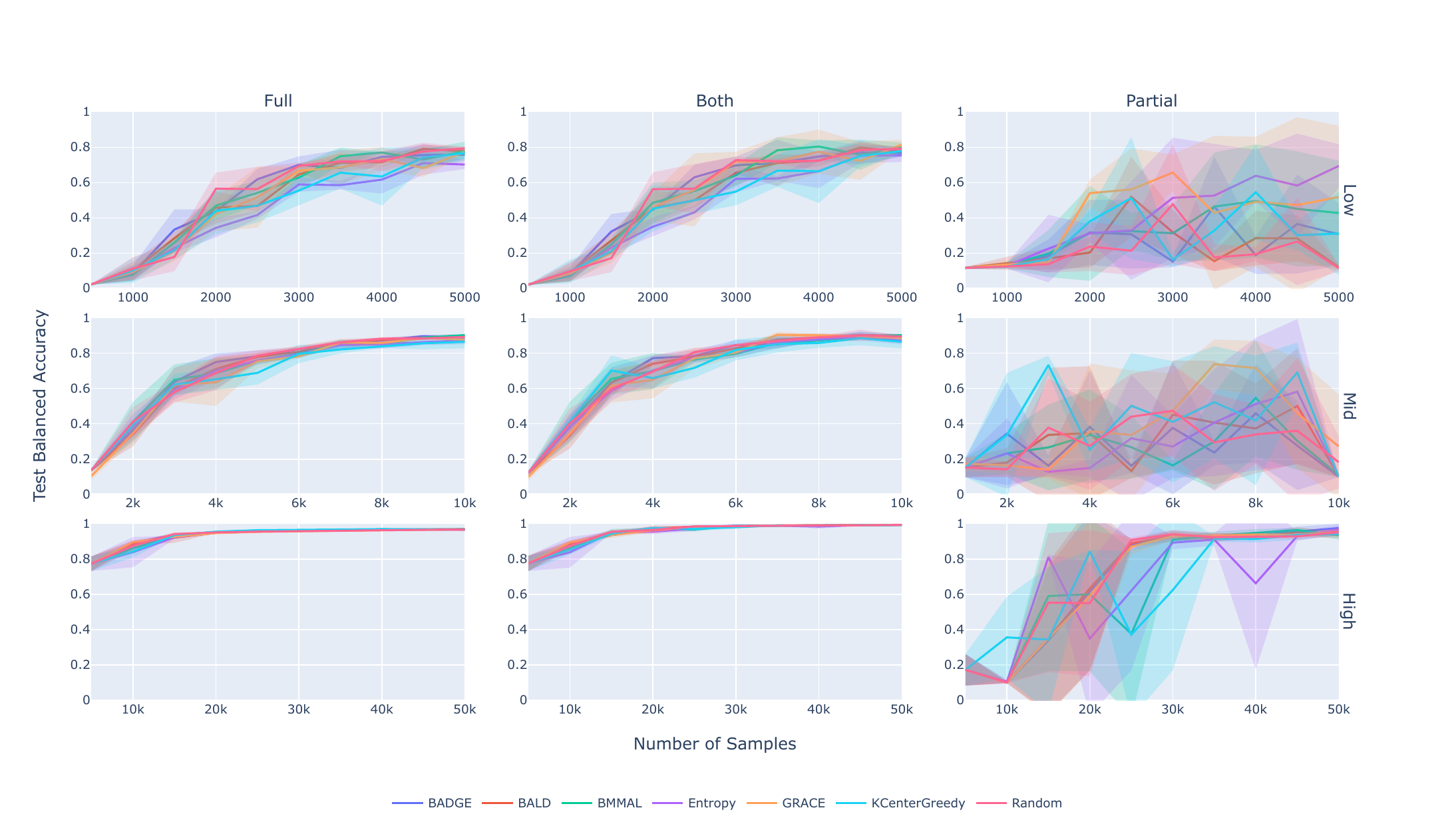}
        \end{subfigure}
        \begin{subfigure}{\textwidth}
            \includegraphics[width=\textwidth]{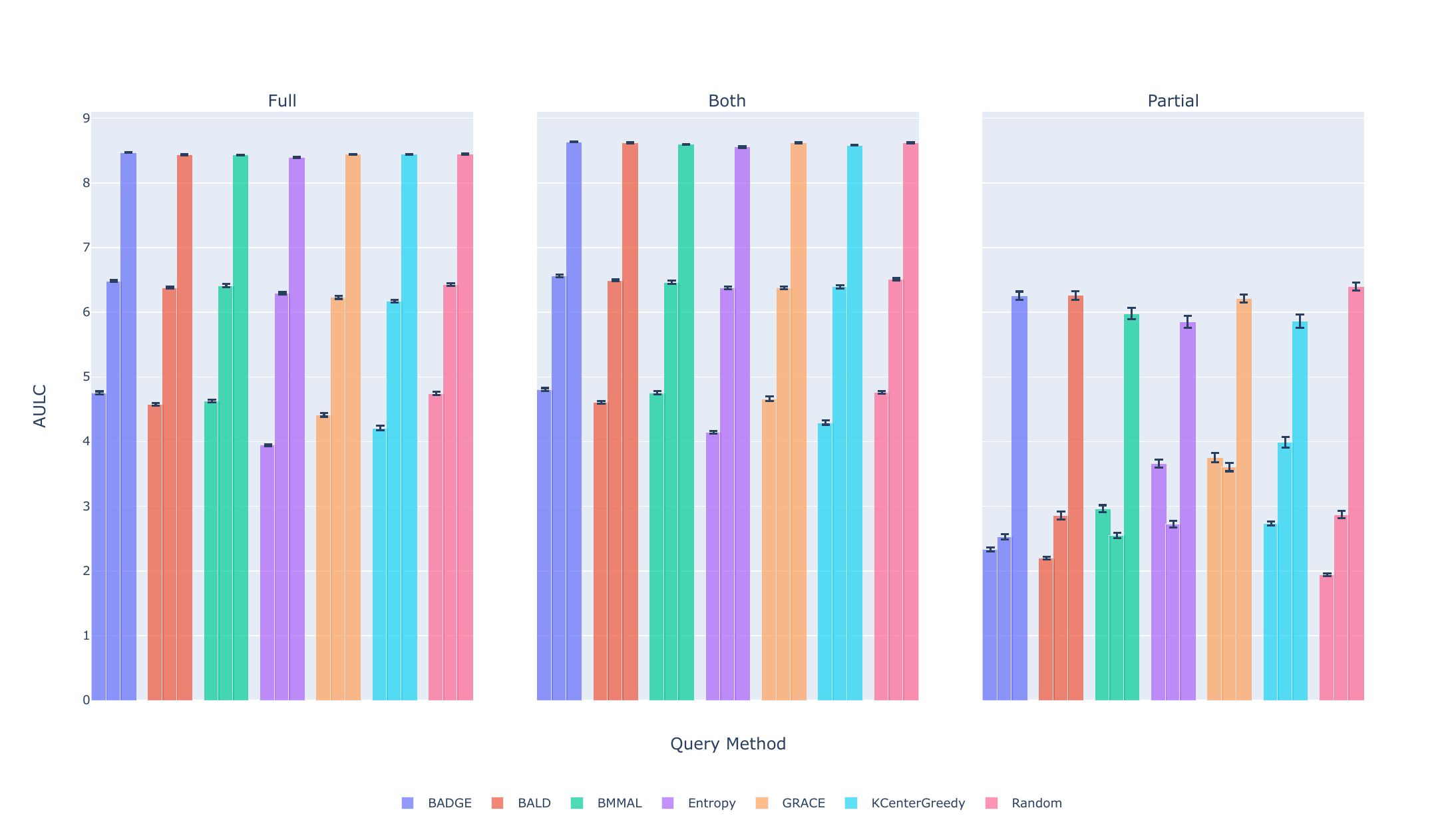}
        \end{subfigure}
    }
    \caption{Results for Unique with ModDrop}\label{fig:results-unique-md}
    \resultsFigureCaption{Unique}
\end{figure}

\begin{table}
\caption{Results for Unique with ModDrop}\label{tab:results-unique-md}
\resultsTableCaption{Unique}
\centering
\begin{tabularx}{\textwidth}{XXXXX}
\toprule
\multicolumn{1}{l}{\textbf{Regime}} & \multicolumn{1}{l}{\textbf{QM}} & \multicolumn{1}{l}{\textbf{A}} & \multicolumn{1}{l}{\textbf{Both}} & \multicolumn{1}{l}{\textbf{B}} \\\midrule\\
Low & BADGE & 4.75 ± 0.48 & 4.80 ± 0.50 & 2.33 ± 1.27 \\
 & BALD & 4.58 ± 0.39 & 4.61 ± 0.43 & 2.20 ± 0.89 \\
 & BMMAL & 4.63 ± 0.43 & 4.76 ± 0.51 & 2.96 ± 1.78 \\
 & Entropy & 3.95 ± 0.36 & 4.14 ± 0.40 & 3.66 ± 1.67 \\
 & GRACE & 4.41 ± 0.61 & 4.66 ± 0.83 & 3.75 ± 1.95 \\
 & KCenterGreedy & 4.21 ± 0.76 & 4.29 ± 0.74 & 2.74 ± 1.09 \\
 & Random & 4.74 ± 0.51 & 4.76 ± 0.50 & 1.94 ± 1.04 \\ \\\midrule \\
Mid & BADGE & 6.48 ± 0.26 & 6.56 ± 0.29 & 2.53 ± 1.72 \\
 & BALD & 6.38 ± 0.26 & 6.49 ± 0.26 & 2.86 ± 2.15 \\
 & BMMAL & 6.41 ± 0.38 & 6.46 ± 0.40 & 2.55 ± 1.63 \\
 & Entropy & 6.29 ± 0.29 & 6.38 ± 0.33 & 2.73 ± 1.92 \\
 & GRACE & 6.23 ± 0.38 & 6.38 ± 0.37 & 3.61 ± 1.84 \\
 & KCenterGreedy & 6.17 ± 0.37 & 6.39 ± 0.43 & 3.99 ± 2.05 \\
 & Random & 6.43 ± 0.32 & 6.52 ± 0.30 & 2.87 ± 2.03 \\ \\\midrule \\
High & BADGE & 8.47 ± 0.05 & 8.64 ± 0.07 & 6.26 ± 1.04 \\
 & BALD & 8.44 ± 0.07 & 8.62 ± 0.06 & 6.26 ± 1.06 \\
 & BMMAL & 8.43 ± 0.05 & 8.60 ± 0.08 & 5.98 ± 1.47 \\
 & Entropy & 8.40 ± 0.13 & 8.56 ± 0.15 & 5.85 ± 1.55 \\
 & GRACE & 8.45 ± 0.05 & 8.62 ± 0.07 & 6.22 ± 1.01 \\
 & KCenterGreedy & 8.44 ± 0.07 & 8.59 ± 0.09 & 5.86 ± 1.69 \\
 & Random & 8.45 ± 0.07 & 8.62 ± 0.09 & 6.40 ± 0.96 \\\\\bottomrule\\
\end{tabularx}
\end{table}

\begin{figure}
    {
        \centering
        \begin{subfigure}{\textwidth}
            \includegraphics[width=\textwidth]{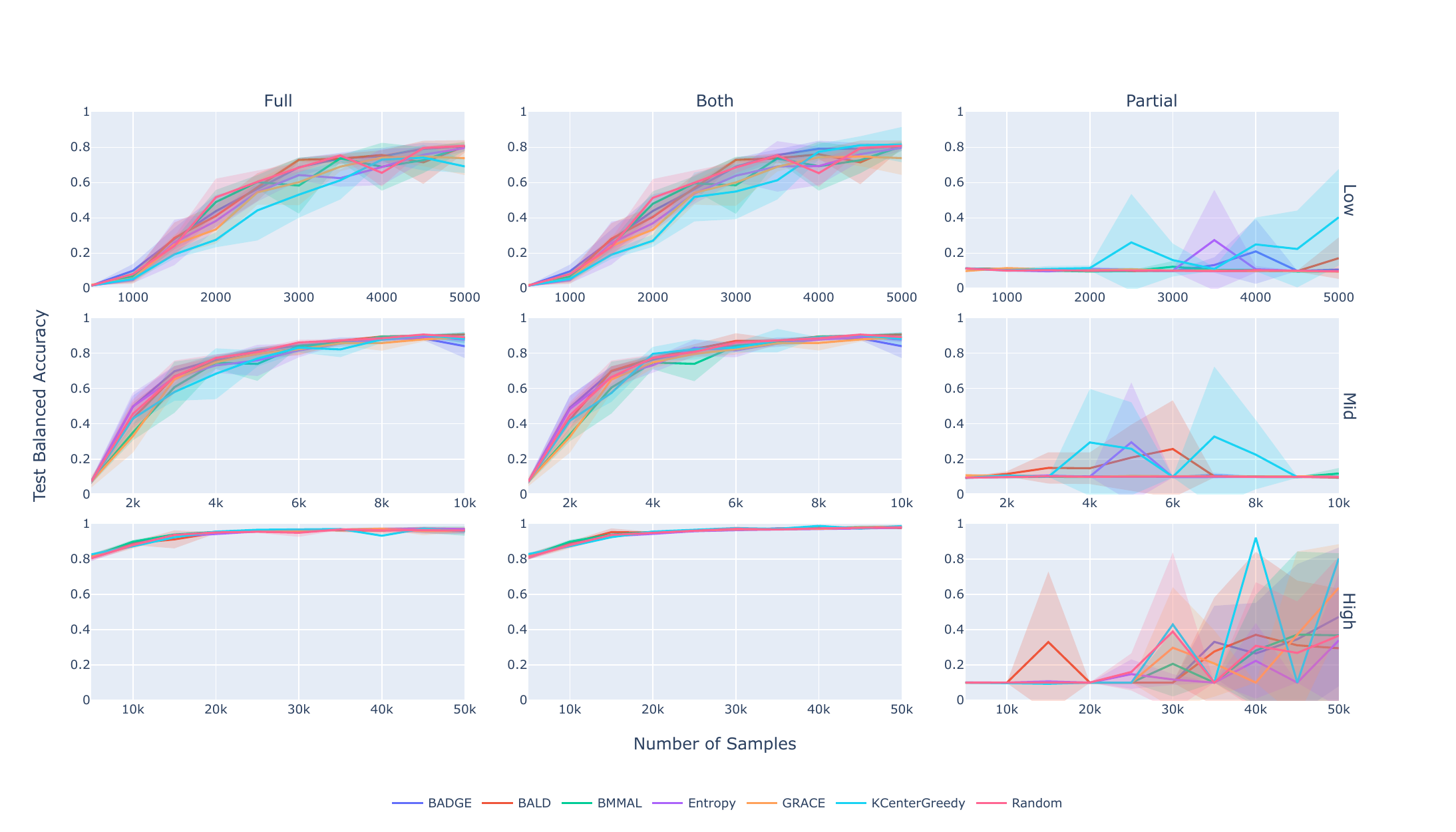}
        \end{subfigure}
        \begin{subfigure}{\textwidth}
            \includegraphics[width=\textwidth]{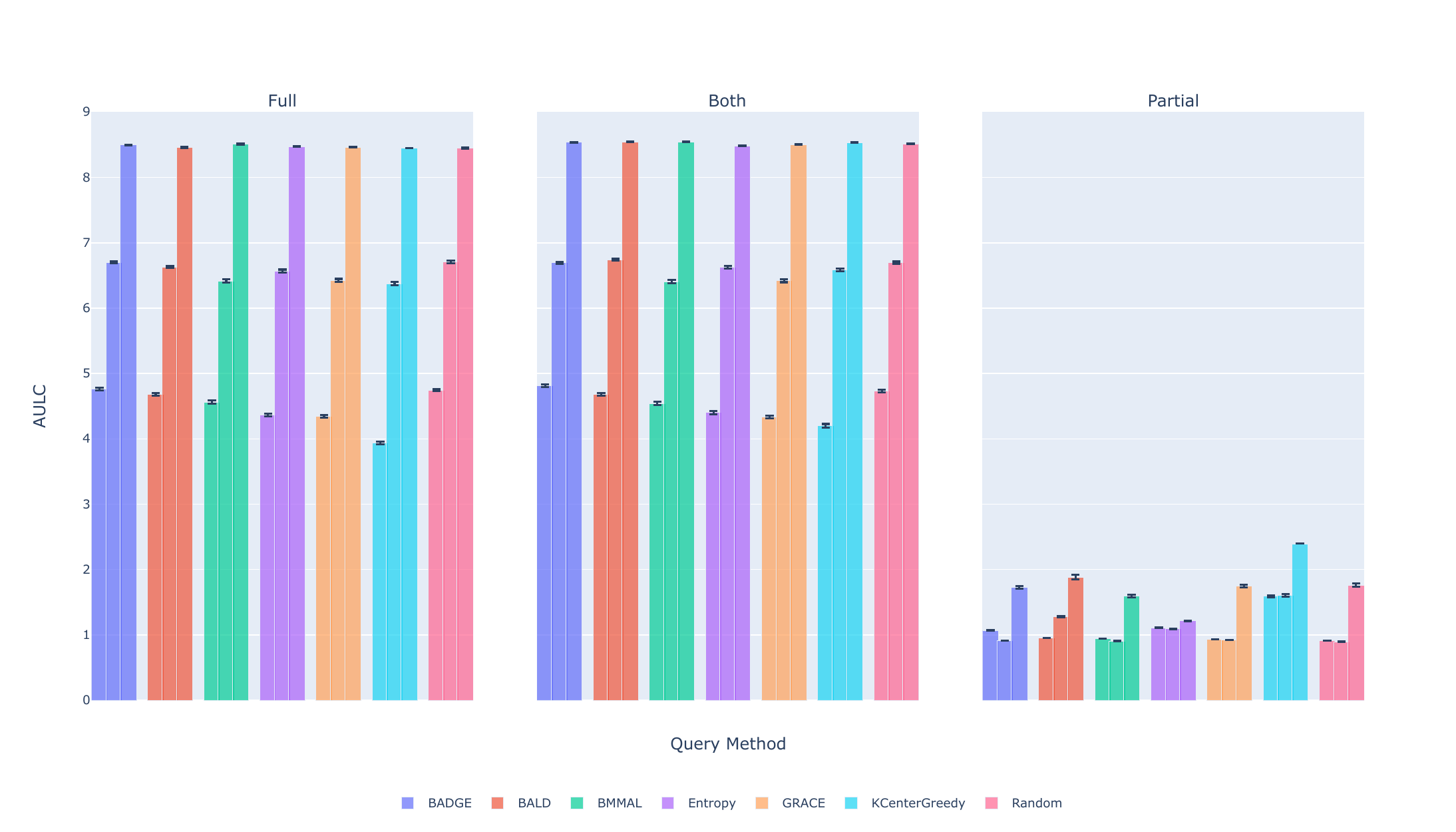}
        \end{subfigure}
    }
    \caption{Results for Unique without ModDrop}\label{fig:results-unique}
    \resultsFigureCaption{Unique}
\end{figure}

\begin{table}
\caption{Results for Unique without ModDrop}\label{tab:results-unique}
\resultsTableCaption{Unique}
\centering
\begin{tabularx}{\textwidth}{XXXXX}
\toprule
\multicolumn{1}{l}{\textbf{Regime}} & \multicolumn{1}{l}{\textbf{QM}} & \multicolumn{1}{l}{\textbf{A}} & \multicolumn{1}{l}{\textbf{Both}} & \multicolumn{1}{l}{\textbf{B}} \\\midrule\\
Low & BADGE & 4.76 ± 0.39 & 4.81 ± 0.38 & 1.07 ± 0.25 \\
 & BALD & 4.68 ± 0.46 & 4.68 ± 0.46 & 0.95 ± 0.08 \\
 & BMMAL & 4.56 ± 0.56 & 4.54 ± 0.56 & 0.94 ± 0.07 \\
 & Entropy & 4.36 ± 0.48 & 4.40 ± 0.58 & 1.11 ± 0.35 \\
 & GRACE & 4.35 ± 0.51 & 4.33 ± 0.51 & 0.94 ± 0.04 \\
 & KCenterGreedy & 3.94 ± 0.59 & 4.20 ± 0.62 & 1.59 ± 0.92 \\
 & Random & 4.75 ± 0.36 & 4.73 ± 0.37 & 0.92 ± 0.03 \\ \\\midrule \\
Mid & BADGE & 6.70 ± 0.23 & 6.70 ± 0.23 & 0.91 ± 0.04 \\
 & BALD & 6.63 ± 0.24 & 6.74 ± 0.23 & 1.28 ± 0.67 \\
 & BMMAL & 6.41 ± 0.37 & 6.41 ± 0.37 & 0.91 ± 0.05 \\
 & Entropy & 6.57 ± 0.36 & 6.63 ± 0.30 & 1.09 ± 0.37 \\
 & GRACE & 6.43 ± 0.34 & 6.42 ± 0.33 & 0.92 ± 0.05 \\
 & KCenterGreedy & 6.38 ± 0.40 & 6.58 ± 0.30 & 1.61 ± 1.17 \\
 & Random & 6.71 ± 0.27 & 6.70 ± 0.26 & 0.90 ± 0.03 \\ \\\midrule \\
High & BADGE & 8.50 ± 0.05 & 8.54 ± 0.04 & 1.73 ± 1.11 \\
 & BALD & 8.46 ± 0.12 & 8.54 ± 0.07 & 1.89 ± 1.71 \\
 & BMMAL & 8.51 ± 0.07 & 8.54 ± 0.06 & 1.59 ± 1.20 \\
 & Entropy & 8.47 ± 0.05 & 8.48 ± 0.05 & 1.22 ± 0.53 \\
 & GRACE & 8.46 ± 0.07 & 8.51 ± 0.06 & 1.74 ± 1.13 \\
 & KCenterGreedy & 8.45 ± 0.00 & 8.53 ± 0.00 & 2.39 ± 0.00 \\
 & Random & 8.45 ± 0.10 & 8.51 ± 0.06 & 1.76 ± 1.43 \\\\\bottomrule\\
\end{tabularx}
\end{table}

\clearpage
\subsection{Synergy}
\begin{figure}
    {
        \centering
        \begin{subfigure}{\textwidth}
            \includegraphics[width=\textwidth]{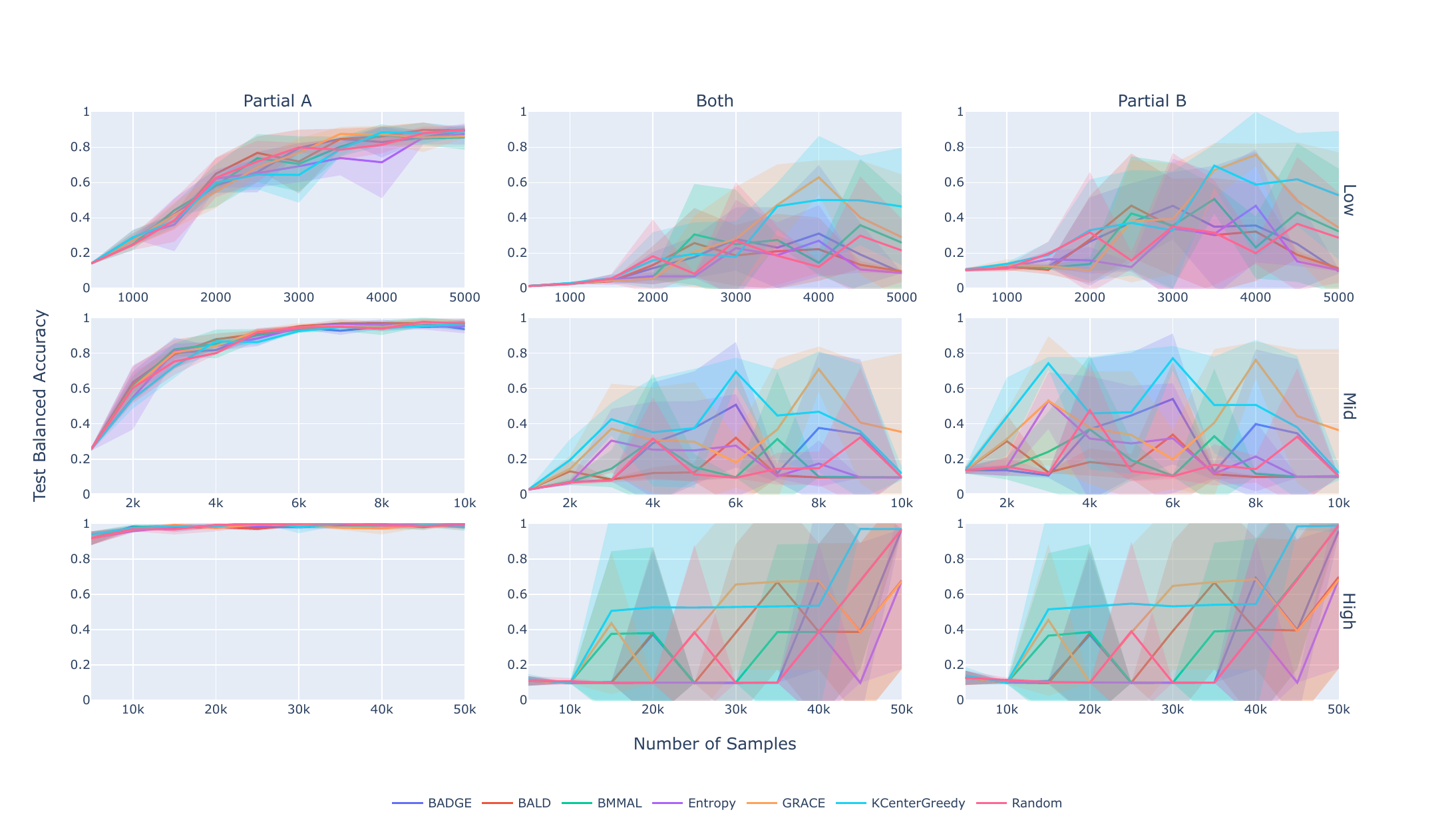}
        \end{subfigure}
        \begin{subfigure}{\textwidth}
            \includegraphics[width=\textwidth]{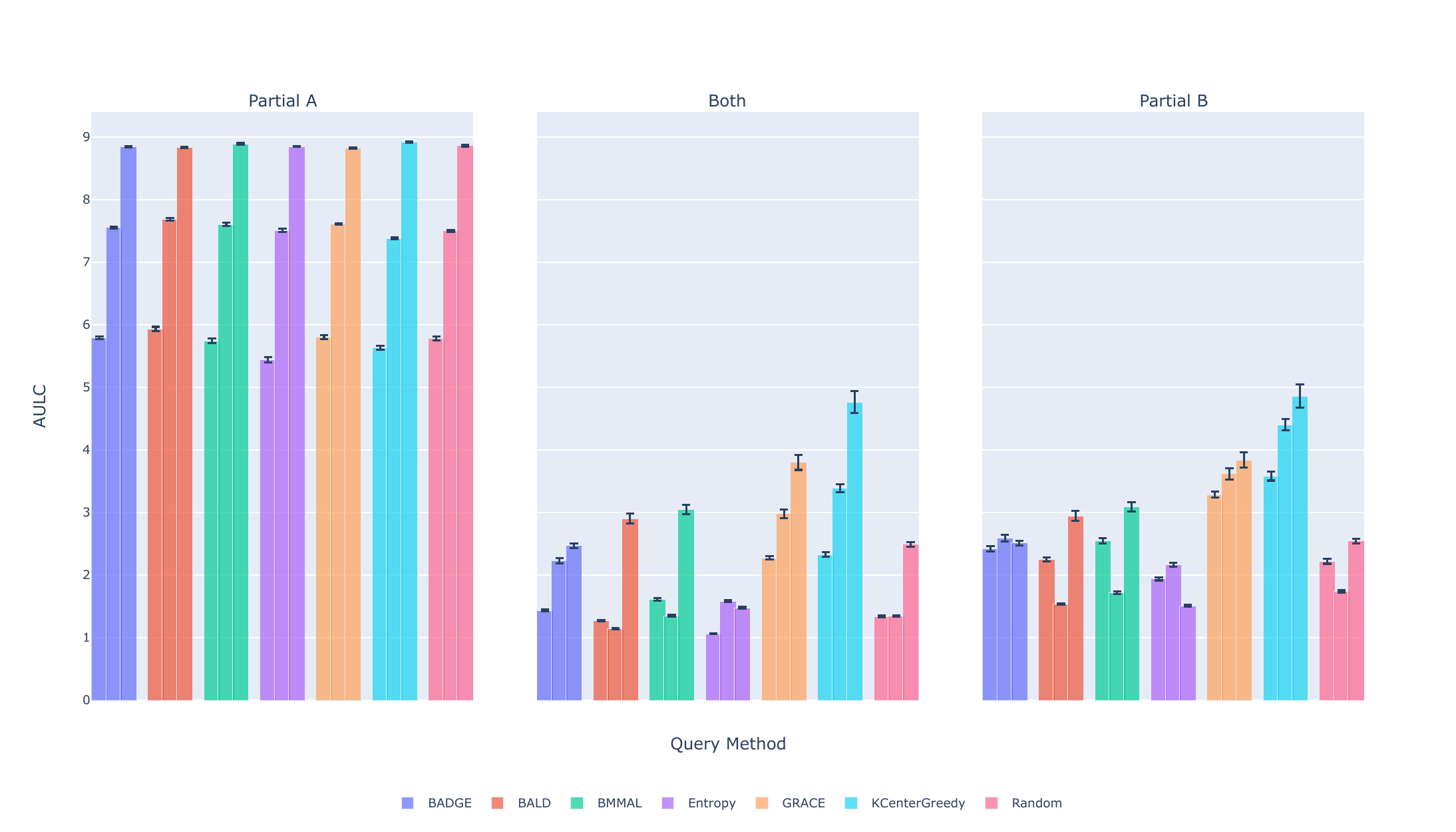}
        \end{subfigure}
    }
    \caption{Results for Synergy with ModDrop}\label{fig:results-synergy-md}
    \resultsFigureCaption{Synergy}
\end{figure}

\begin{table}
\caption{Results for Synergy with ModDrop}\label{tab:results-synergy-md}
\resultsTableCaption{Synergy}
\centering
\begin{tabularx}{\textwidth}{XXXXX}
\toprule
\multicolumn{1}{l}{\textbf{Regime}} & \multicolumn{1}{l}{\textbf{QM}} & \multicolumn{1}{l}{\textbf{A}} & \multicolumn{1}{l}{\textbf{Both}} & \multicolumn{1}{l}{\textbf{B}} \\\midrule\\
Low & BADGE & 5.79 ± 0.35 & 1.43 ± 1.19 & 2.42 ± 1.73 \\
 & BALD & 5.93 ± 0.62 & 1.27 ± 0.91 & 2.25 ± 1.49 \\
 & BMMAL & 5.74 ± 0.67 & 1.61 ± 1.22 & 2.55 ± 1.57 \\
 & Entropy & 5.44 ± 0.75 & 1.06 ± 0.74 & 1.94 ± 1.26 \\
 & GRACE & 5.80 ± 0.51 & 2.28 ± 1.34 & 3.29 ± 1.48 \\
 & KCenterGreedy & 5.63 ± 0.52 & 2.33 ± 1.48 & 3.58 ± 1.99 \\
 & Random & 5.78 ± 0.56 & 1.34 ± 1.19 & 2.22 ± 1.76 \\ \\\midrule \\
Mid & BADGE & 7.55 ± 0.25 & 2.23 ± 1.93 & 2.59 ± 2.07 \\
 & BALD & 7.68 ± 0.32 & 1.14 ± 0.57 & 1.54 ± 0.76 \\
 & BMMAL & 7.60 ± 0.34 & 1.35 ± 0.96 & 1.72 ± 1.24 \\
 & Entropy & 7.51 ± 0.38 & 1.58 ± 1.18 & 2.16 ± 1.44 \\
 & GRACE & 7.61 ± 0.20 & 2.98 ± 2.20 & 3.62 ± 2.51 \\
 & KCenterGreedy & 7.38 ± 0.22 & 3.39 ± 1.93 & 4.40 ± 2.07 \\
 & Random & 7.50 ± 0.23 & 1.35 ± 0.88 & 1.74 ± 0.99 \\ \\\midrule \\
High & BADGE & 8.84 ± 0.12 & 2.47 ± 1.50 & 2.51 ± 1.55 \\
 & BALD & 8.83 ± 0.11 & 2.90 ± 2.75 & 2.95 ± 2.81 \\
 & BMMAL & 8.89 ± 0.15 & 3.05 ± 2.47 & 3.09 ± 2.52 \\
 & Entropy & 8.85 ± 0.10 & 1.48 ± 0.78 & 1.51 ± 0.80 \\
 & GRACE & 8.82 ± 0.13 & 3.80 ± 3.14 & 3.84 ± 3.18 \\
 & KCenterGreedy & 8.92 ± 0.11 & 4.77 ± 3.64 & 4.86 ± 3.76 \\
 & Random & 8.86 ± 0.17 & 2.49 ± 1.54 & 2.54 ± 1.58 \\\\\bottomrule\\
\end{tabularx}
\end{table}

\begin{figure}
    {
        \centering
        \begin{subfigure}{\textwidth}
            \includegraphics[width=\textwidth]{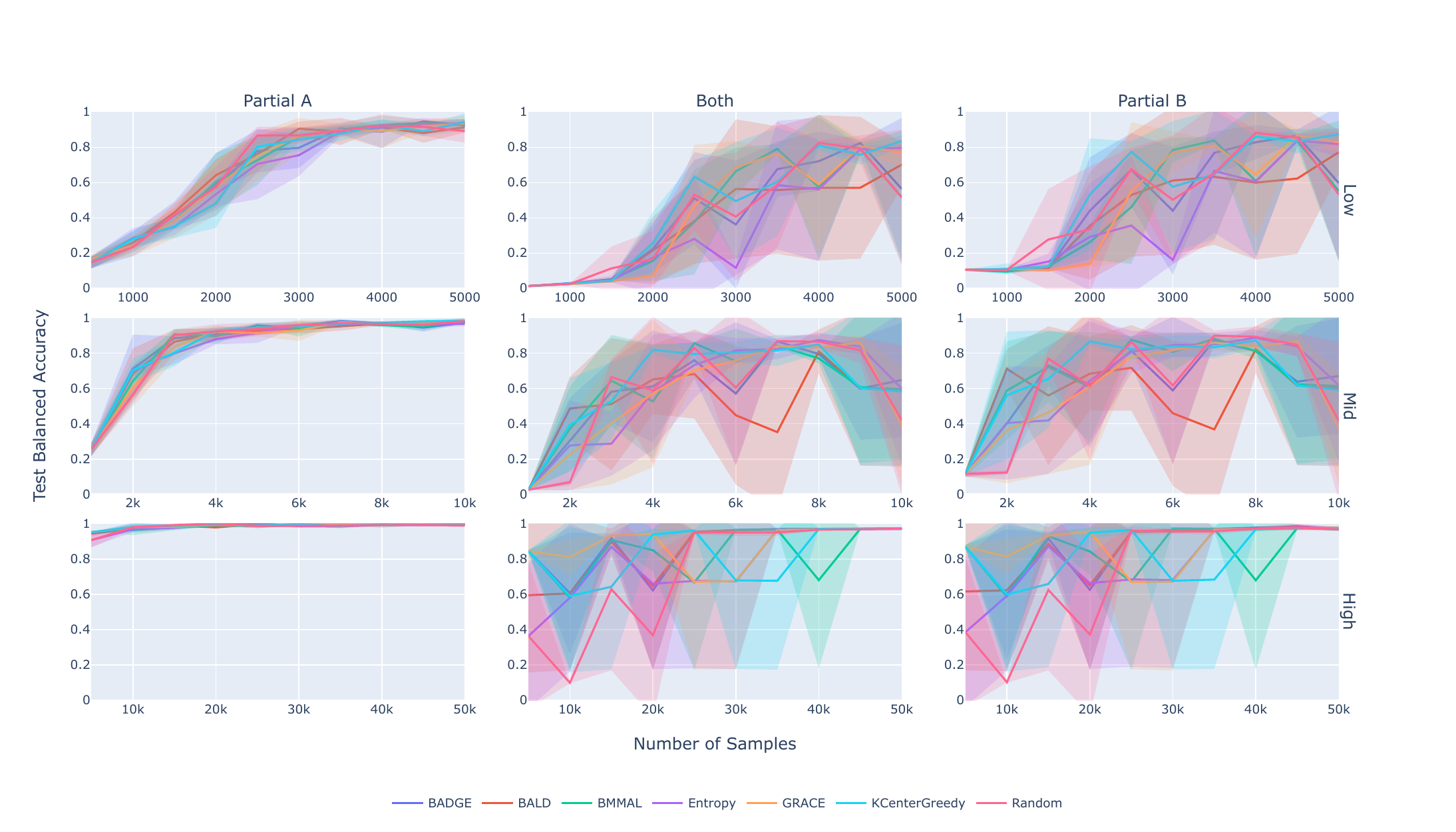}
        \end{subfigure}
        \begin{subfigure}{\textwidth}
            \includegraphics[width=\textwidth]{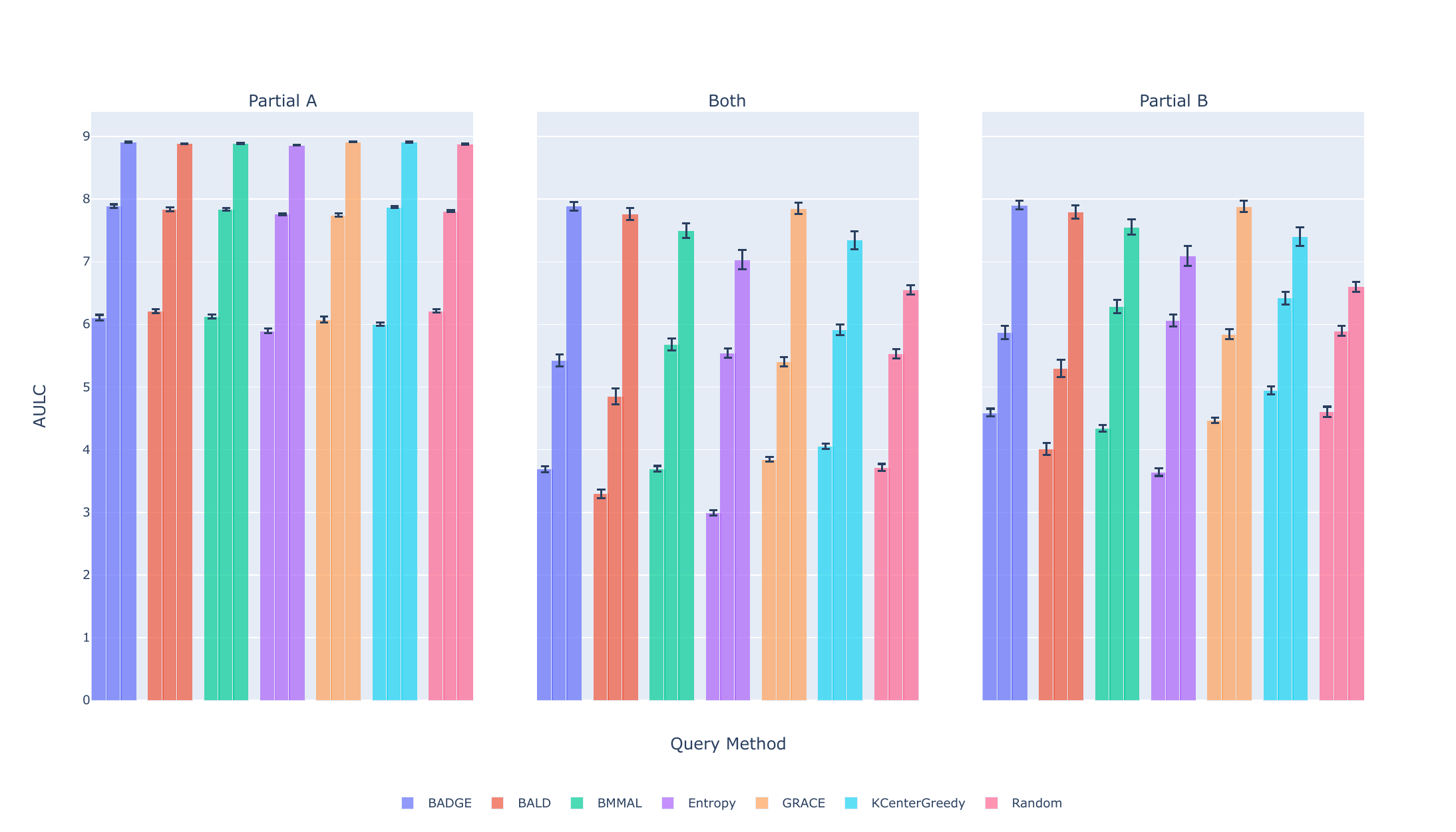}
        \end{subfigure}
    }
    \caption{Results for Synergy without ModDrop}\label{fig:results-synergy}
    \resultsFigureCaption{Synergy}
\end{figure}

\begin{table}
\caption{Results for Synergy without ModDrop}\label{tab:results-synergy}
\resultsTableCaption{Synergy}
\centering
\begin{tabularx}{\textwidth}{XXXXX}
\toprule
\multicolumn{1}{l}{\textbf{Regime}} & \multicolumn{1}{l}{\textbf{QM}} & \multicolumn{1}{l}{\textbf{A}} & \multicolumn{1}{l}{\textbf{Both}} & \multicolumn{1}{l}{\textbf{B}} \\\midrule\\
Low & BADGE & 6.11 ± 0.72 & 3.69 ± 1.33 & 4.59 ± 1.38 \\
 & BALD & 6.21 ± 0.46 & 3.30 ± 2.07 & 4.01 ± 2.40 \\
 & BMMAL & 6.13 ± 0.51 & 3.70 ± 1.23 & 4.34 ± 1.20 \\
 & Entropy & 5.90 ± 0.59 & 2.99 ± 1.41 & 3.64 ± 1.62 \\
 & GRACE & 6.08 ± 0.76 & 3.85 ± 1.02 & 4.47 ± 0.99 \\
 & KCenterGreedy & 6.00 ± 0.51 & 4.06 ± 1.02 & 4.95 ± 1.25 \\
 & Random & 6.22 ± 0.43 & 3.72 ± 1.49 & 4.61 ± 1.74 \\ \\\midrule \\
Mid & BADGE & 7.89 ± 0.35 & 5.42 ± 1.74 & 5.87 ± 1.77 \\
 & BALD & 7.84 ± 0.36 & 4.85 ± 2.62 & 5.30 ± 2.65 \\
 & BMMAL & 7.84 ± 0.27 & 5.68 ± 1.68 & 6.29 ± 1.67 \\
 & Entropy & 7.76 ± 0.24 & 5.54 ± 1.38 & 6.06 ± 1.56 \\
 & GRACE & 7.74 ± 0.35 & 5.40 ± 1.34 & 5.84 ± 1.44 \\
 & KCenterGreedy & 7.87 ± 0.25 & 5.91 ± 1.42 & 6.42 ± 1.56 \\
 & Random & 7.81 ± 0.21 & 5.53 ± 1.36 & 5.90 ± 1.34 \\ \\\midrule \\
High & BADGE & 8.91 ± 0.10 & 7.89 ± 0.85 & 7.90 ± 0.91 \\
 & BALD & 8.89 ± 0.04 & 7.76 ± 1.22 & 7.79 ± 1.32 \\
 & BMMAL & 8.89 ± 0.10 & 7.50 ± 1.54 & 7.55 ± 1.60 \\
 & Entropy & 8.86 ± 0.09 & 7.04 ± 2.14 & 7.09 ± 2.21 \\
 & GRACE & 8.92 ± 0.05 & 7.85 ± 1.13 & 7.89 ± 1.16 \\
 & KCenterGreedy & 8.91 ± 0.07 & 7.35 ± 1.95 & 7.40 ± 2.00 \\
 & Random & 8.88 ± 0.09 & 6.55 ± 1.19 & 6.60 ± 1.22 \\\\\bottomrule\\
\end{tabularx}
\end{table}

\clearpage
\subsection{Food101}
\begin{figure}
    {
        \centering
        \begin{subfigure}{\textwidth}
            \includegraphics[width=\textwidth]{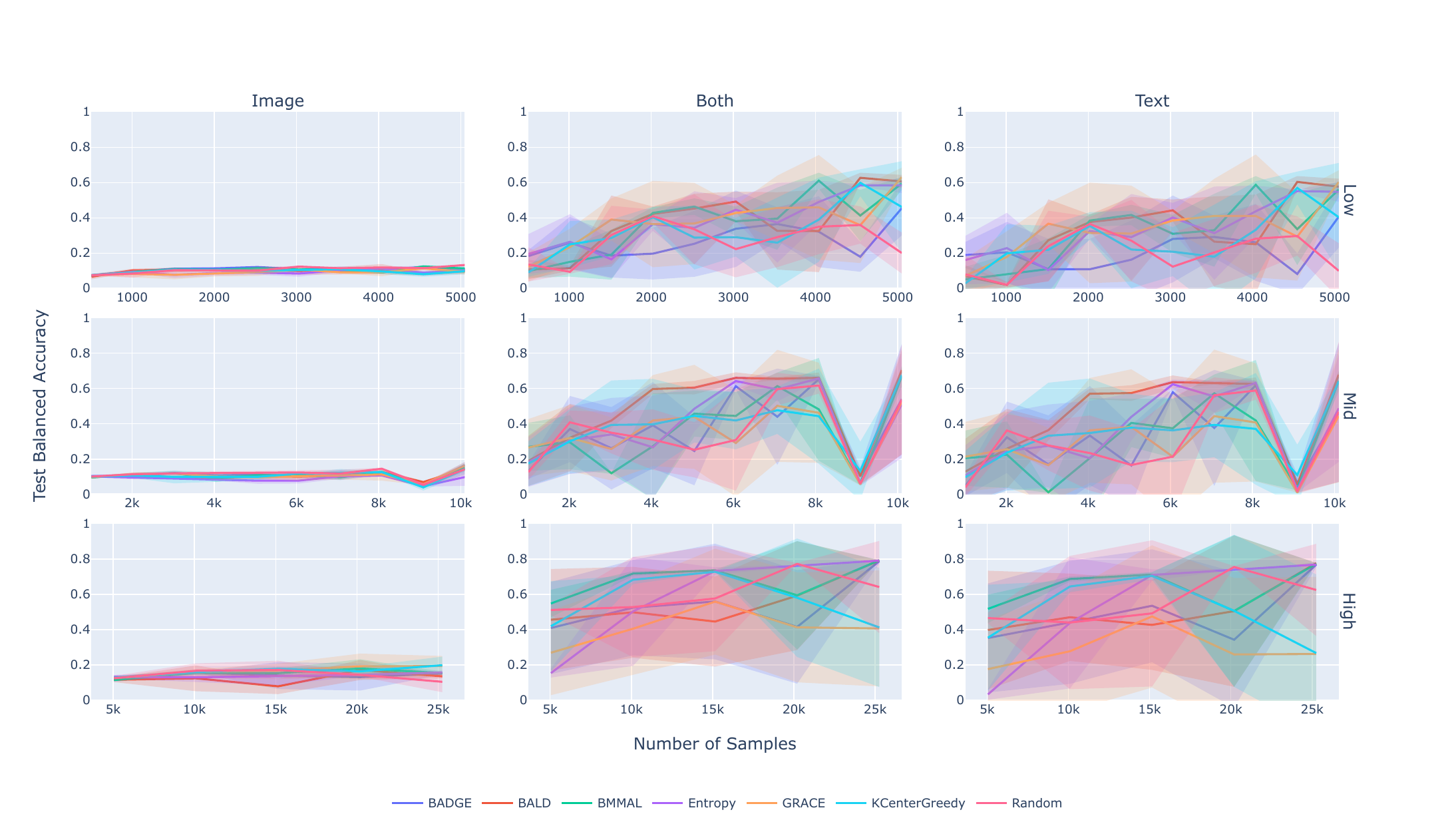}
        \end{subfigure}
        \begin{subfigure}{\textwidth}
            \includegraphics[width=\textwidth]{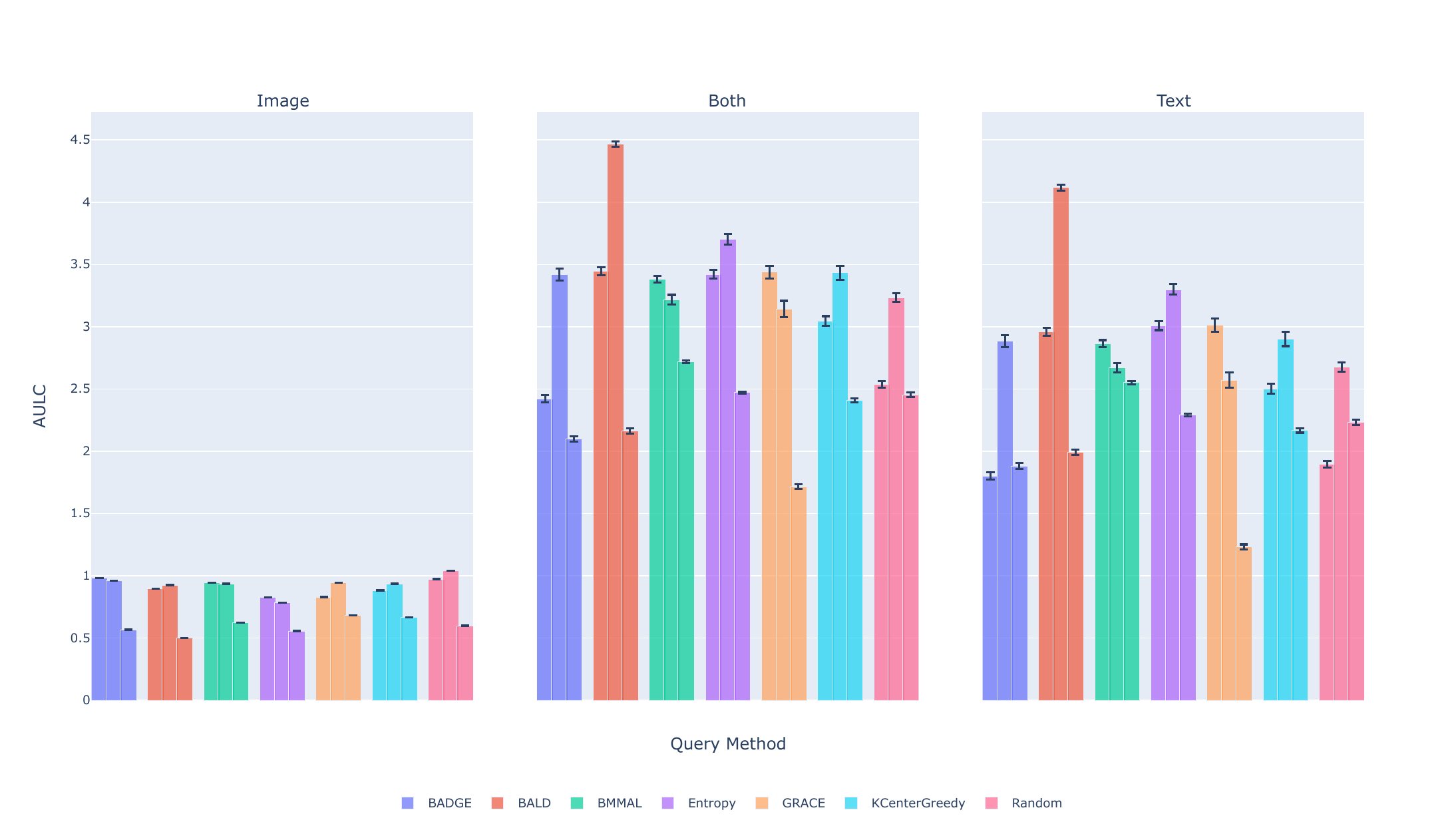}
        \end{subfigure}
    }
    \caption{Results for Food101 with ModDrop}\label{fig:results-food101}
    \resultsFigureCaption{Food101}
\end{figure}

\begin{table}
\caption{Results for Food101 with ModDrop}\label{tab:results-food101-md}
\resultsTableCaption{Food101}
\centering
\begin{tabularx}{\textwidth}{XXXXX}
\toprule
\multicolumn{1}{l}{\textbf{Regime}} & \multicolumn{1}{l}{\textbf{QM}} & \multicolumn{1}{l}{\textbf{Image}} & \multicolumn{1}{l}{\textbf{Both}} & \multicolumn{1}{l}{\textbf{Text}} \\\midrule\\
Low & BADGE & 0.98 ± 0.08 & 2.42 ± 1.30 & 1.80 ± 1.56 \\
 & BALD & 0.90 ± 0.14 & 3.45 ± 0.93 & 2.96 ± 1.10 \\
 & BMMAL & 0.94 ± 0.07 & 3.38 ± 0.79 & 2.86 ± 1.00 \\
 & Entropy & 0.83 ± 0.11 & 3.42 ± 1.00 & 3.01 ± 1.21 \\
 & GRACE & 0.83 ± 0.14 & 3.44 ± 1.51 & 3.01 ± 1.77 \\
 & KCenterGreedy & 0.88 ± 0.09 & 3.05 ± 1.30 & 2.50 ± 1.57 \\
 & Random & 0.97 ± 0.08 & 2.54 ± 1.11 & 1.90 ± 1.30 \\ \\\midrule \\
Mid & BADGE & 0.96 ± 0.10 & 3.42 ± 1.40 & 2.88 ± 1.64 \\
 & BALD & 0.92 ± 0.09 & 4.47 ± 0.48 & 4.12 ± 0.58 \\
 & BMMAL & 0.93 ± 0.09 & 3.22 ± 1.19 & 2.67 ± 1.40 \\
 & Entropy & 0.78 ± 0.10 & 3.70 ± 1.16 & 3.30 ± 1.30 \\
 & GRACE & 0.94 ± 0.16 & 3.14 ± 2.06 & 2.57 ± 2.40 \\
 & KCenterGreedy & 0.93 ± 0.16 & 3.43 ± 1.65 & 2.90 ± 1.96 \\
 & Random & 1.04 ± 0.10 & 3.23 ± 1.12 & 2.68 ± 1.38 \\ \\\midrule \\
High & BADGE & 0.57 ± 0.17 & 2.10 ± 1.06 & 1.88 ± 1.21 \\
 & BALD & 0.50 ± 0.19 & 2.16 ± 0.96 & 1.99 ± 1.10 \\
 & BMMAL & 0.62 ± 0.10 & 2.72 ± 0.36 & 2.55 ± 0.50 \\
 & Entropy & 0.56 ± 0.10 & 2.47 ± 0.35 & 2.29 ± 0.42 \\
 & GRACE & 0.68 ± 0.16 & 1.72 ± 1.16 & 1.23 ± 1.54 \\
 & KCenterGreedy & 0.67 ± 0.08 & 2.41 ± 0.66 & 2.17 ± 0.86 \\
 & Random & 0.60 ± 0.14 & 2.45 ± 0.73 & 2.23 ± 0.94 \\\\\bottomrule\\
\end{tabularx}
\end{table}

\clearpage
\subsection{MIMIC}
\begin{figure}
    {
        \centering
        \begin{subfigure}{\textwidth}
            \includegraphics[width=\textwidth]{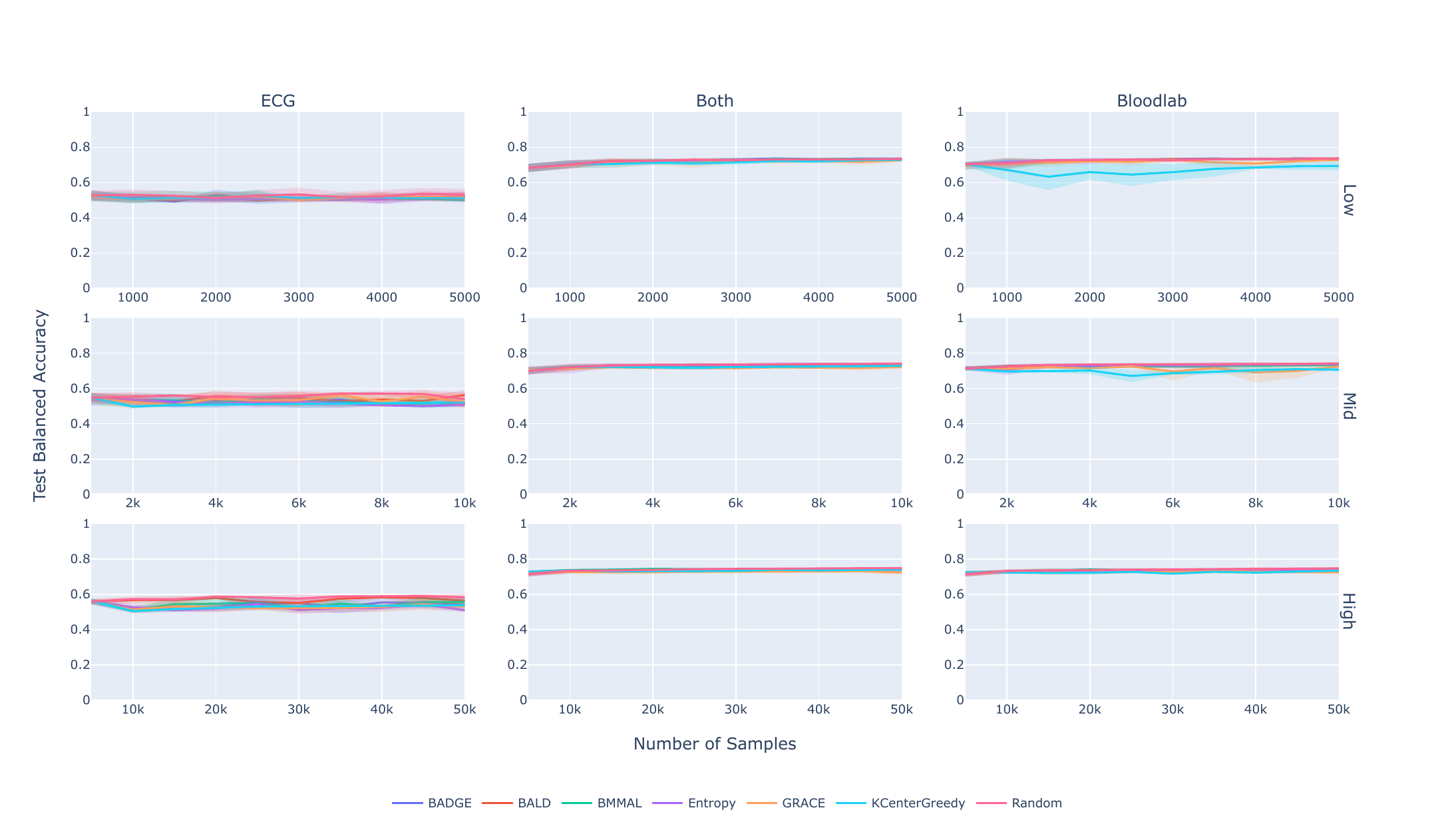}
        \end{subfigure}
        \begin{subfigure}{\textwidth}
            \includegraphics[width=\textwidth]{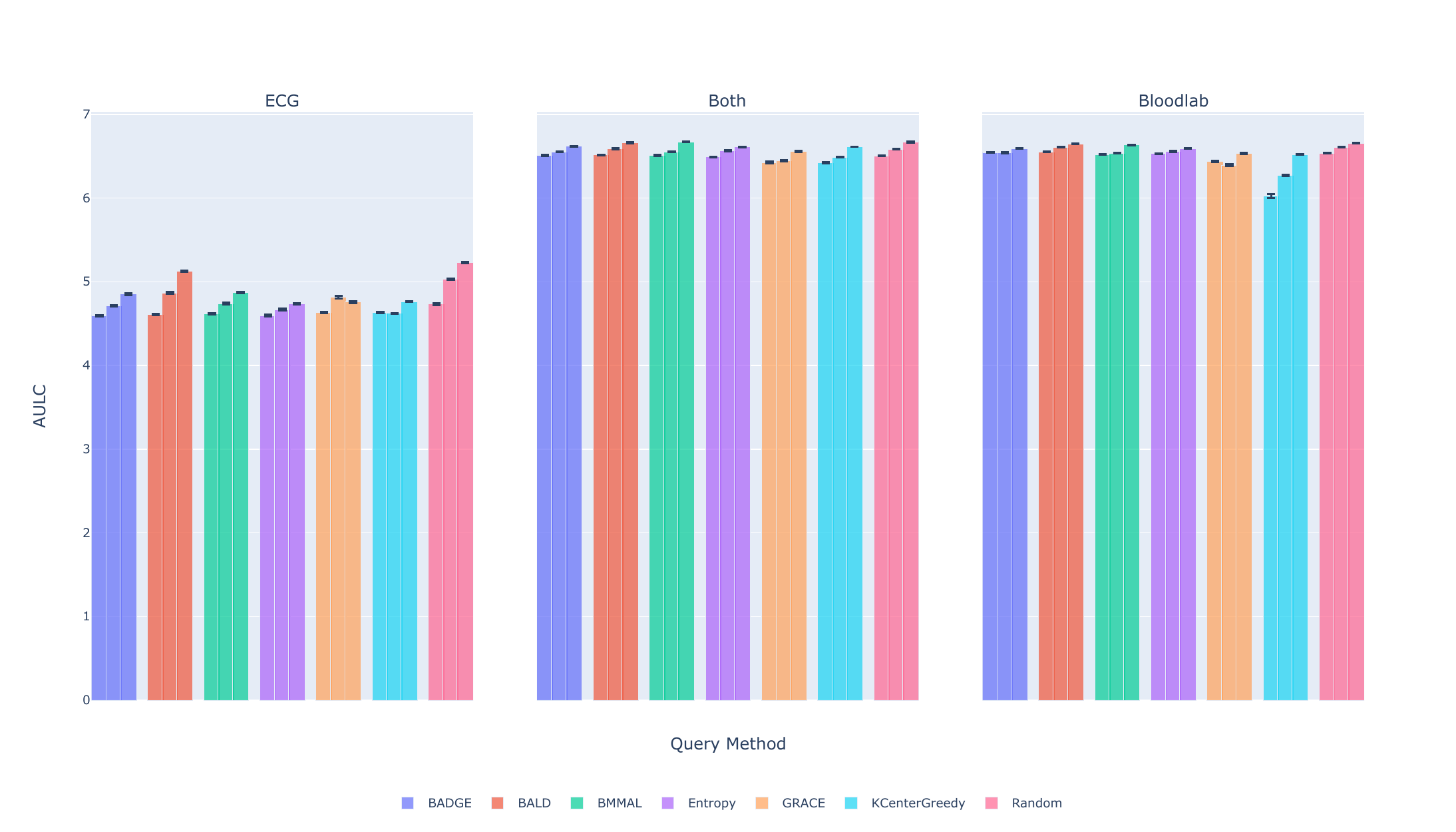}
        \end{subfigure}
    }
    \caption{Results for MIMIC with ModDrop}\label{fig:results-mimic}
    \resultsFigureCaption{MIMIC}
\end{figure}

\begin{table}
\caption{Results for MIMIC with ModDrop}\label{tab:results-mimic-md}
\resultsTableCaption{MIMIC}
\centering
\begin{tabularx}{\textwidth}{XXXXX}
\toprule
\multicolumn{1}{l}{\textbf{Regime}} & \multicolumn{1}{l}{\textbf{QM}} & \multicolumn{1}{l}{\textbf{ECG}} & \multicolumn{1}{l}{\textbf{Both}} & \multicolumn{1}{l}{\textbf{Bloodlabs}} \\\midrule\\
Low & BADGE & 4.60 ± 0.18 & 6.51 ± 0.10 & 6.55 ± 0.08 \\
 & BALD & 4.61 ± 0.17 & 6.52 ± 0.08 & 6.55 ± 0.07 \\
 & BMMAL & 4.62 ± 0.17 & 6.51 ± 0.10 & 6.52 ± 0.10 \\
 & Entropy & 4.60 ± 0.19 & 6.49 ± 0.08 & 6.53 ± 0.08 \\
 & GRACE & 4.63 ± 0.18 & 6.43 ± 0.16 & 6.44 ± 0.16 \\
 & KCenterGreedy & 4.64 ± 0.16 & 6.42 ± 0.11 & 6.03 ± 0.38 \\
 & Random & 4.73 ± 0.28 & 6.51 ± 0.08 & 6.54 ± 0.08 \\ \\\midrule \\
Mid & BADGE & 4.71 ± 0.20 & 6.55 ± 0.08 & 6.54 ± 0.09 \\
 & BALD & 4.87 ± 0.28 & 6.59 ± 0.08 & 6.61 ± 0.06 \\
 & BMMAL & 4.74 ± 0.27 & 6.55 ± 0.07 & 6.54 ± 0.09 \\
 & Entropy & 4.67 ± 0.20 & 6.57 ± 0.11 & 6.56 ± 0.13 \\
 & GRACE & 4.82 ± 0.26 & 6.45 ± 0.09 & 6.40 ± 0.20 \\
 & KCenterGreedy & 4.62 ± 0.07 & 6.49 ± 0.06 & 6.27 ± 0.14 \\
 & Random & 5.03 ± 0.19 & 6.58 ± 0.06 & 6.61 ± 0.05 \\ \\\midrule \\
High & BADGE & 4.85 ± 0.26 & 6.62 ± 0.08 & 6.59 ± 0.08 \\
 & BALD & 5.13 ± 0.18 & 6.66 ± 0.06 & 6.65 ± 0.06 \\
 & BMMAL & 4.87 ± 0.22 & 6.67 ± 0.05 & 6.64 ± 0.06 \\
 & Entropy & 4.74 ± 0.17 & 6.61 ± 0.07 & 6.59 ± 0.07 \\
 & GRACE & 4.76 ± 0.23 & 6.56 ± 0.10 & 6.53 ± 0.09 \\
 & KCenterGreedy & 4.76 ± 0.05 & 6.61 ± 0.03 & 6.52 ± 0.04 \\
 & Random & 5.23 ± 0.09 & 6.67 ± 0.06 & 6.66 ± 0.06 \\\\\bottomrule\\
\end{tabularx}
\end{table}